\documentclass[10pt,twocolumn,letterpaper]{article}

\usepackage{wacv}

\usepackage{times}
\usepackage{epsfig}
\usepackage{graphicx}
\usepackage{amsmath}
\usepackage{amssymb}
\usepackage{url}
\usepackage{enumitem}
\usepackage{soul}
\usepackage{multirow}
\usepackage{multicol}
\usepackage{booktabs}
\usepackage[dvipsnames]{xcolor}
\usepackage{makecell}
\usepackage[accsupp]{axessibility}  %
\usepackage[title]{appendix}

\def\wacvPaperID{1819} %

\wacvalgorithmstrack   %

\wacvfinalcopy %

\ifwacvfinal
\usepackage[breaklinks=true,bookmarks=false]{hyperref}
\else
\usepackage[pagebackref=true,breaklinks=true,colorlinks,bookmarks=false]{hyperref}
\fi

\definecolor{mygray}{gray}{0.4}
\newcolumntype{x}[1]{>{\centering\arraybackslash}p{#1pt}}
\newlength\savewidth

\makeatletter\renewcommand\paragraph{\@startsection{paragraph}{4}{\z@}
  {.5em \@plus1ex \@minus.2ex}{-.5em}{\normalfont\normalsize\bfseries}}\makeatother

\newcommand{\app}{\raise.17ex\hbox{$\scriptstyle\sim$}}

\newcommand{\new}[1]{#1}

\begin{document}

\title{DCVNet: Dilated Cost Volume Networks for Fast Optical Flow}

\author{Huaizu Jiang\\
Northeastern University\\
Boston, MA 02115\\
{\tt\small h.jiang@northeastern.edu}
\and
Erik Learned-Miller\\
UMass Amherst\\
Amherst, MA 01003\\
{\tt\small elm@cs.umass.edu}
}

\maketitle

\pagestyle{empty}
\thispagestyle{empty}

\begin{abstract}

\new{The cost volume, capturing the similarity of possible correspondences across two input images, is a key ingredient in state-of-the-art optical flow approaches. When sampling correspondences to build the cost volume, a large neighborhood radius is required to deal with large displacements, introducing a significant computational burden. To address this, coarse-to-fine or recurrent processing of the cost volume is usually adopted, where correspondence sampling in a local neighborhood with a small radius suffices. 
}
In this paper, we propose an alternative by constructing cost volumes with different dilation factors to capture small and large displacements simultaneously. 
A U-Net with sikp connections is employed to convert the dilated cost volumes into interpolation weights between all possible captured displacements to get the optical flow. 
Our proposed model DCVNet only needs to process the cost volume once in a simple feedforward manner and does not rely on the sequential processing strategy.
DCVNet obtains comparable accuracy to existing approaches and achieves real-time inference (\new{30} fps on a mid-end 1080ti GPU).
The code and model weights are available at \url{https://github.com/neu-vi/ezflow}.
\end{abstract}

\section{Introduction}
Optical flow, as a dense matching problem, is about estimating every single pixel's displacement between two consecutive video frames, capturing the motion of brightness patterns. It is a classical and long-studied problem in computer vision, dating back to the early 1980s~\cite{Horn:1981:DO}. Optical flow has applications in a wide range of other problems, such as scene flow estimation~\cite{menze2015object}, action recognition~\cite{simonyan14two}, and video editing and synthesis~\cite{baker11a}.

\begin{figure}[t]
\centering
\includegraphics[width=0.98\linewidth]{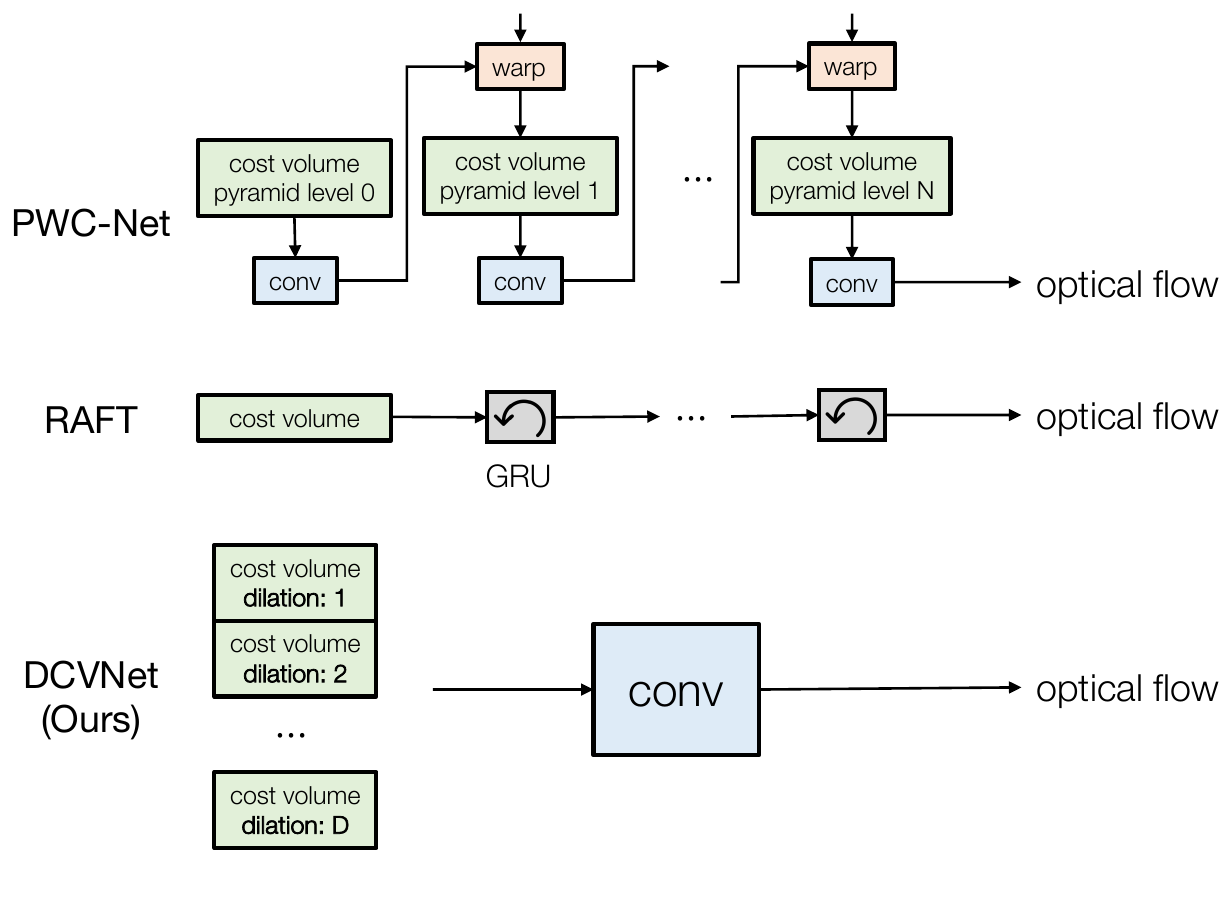}
\caption{\textbf{Illustrations of our proposed model, DCVNet, compared with two representative existing approaches.} DCVNet is an alternative to existing approaches, which does not need the sequential processing of the cost volume. The key idea is to
construct cost volumes with different dilation rates to capture small and large displacement at the same time. It achieves real-time inference on a mid-end 1080ti GPU (30 fps) and comparable accuracy to existing approaches.}
\label{fig:teaser}
\end{figure}

Like many other computer vision problems, state-of-the-art approaches for optical flow estimation are all based on deep neural networks. In the beginning, however, deep neural networks for optical flow reported inferior results compared to those of traditional well-engineered energy minimization  approaches~\cite{Dosovitskiy:2015Flownet}. To reduce the performance gap, one approach stacked multiple networks to increase capacity~\cite{Ilg:2016:Flownet2}, resulting in significant accuracy improvements. Increasing the capacity of the network, however, led to huge networks and slower inference. Starting from~\cite{Ranjan:2016:SpyNet}, more and more classical principles derived from traditional optical flow estimation approaches were incorporated into neural network design, allowing deep neural networks to surpass traditional approaches in accuracy. \new{In particular, the cost volume, which is a more discriminative representation for optical flow  compared to concatenated feature representations of two images, is now an essential component for state-of-the-art approaches.}

\new{To build the cost volume, we need to sample pairs of locations between two input images in a neighborhood along both horizontal and vertical directions to compute their similarity (or cost). A big neighborhood is required to capture large displacements, but it leads to a very large cost volume and a significant computational burden. Consequently, most existing models capture large displacements in a sequential manner by using cost volumes with small neighborhoods for computational efficiency. 
Specifically, the coarse-to-fine strategy is widely adopted in state-of-the-art advances~\cite{sun2018pwc,Hui_2018_CVPR,yang2019volumetric,wang2020displacement}, where a pyramid based on the feature hierarchy of deep Convolutional Neural Networks (CNNs) is built. In each pyramid level, the optical flow estimation in the previous level is used to construct the cost volume with the warping operation. Although a full-range cost volume is constructed in the recent ground-breaking work~\cite{teed2020raft}, a Recurrent Neural Network (RNN) is employed to process only a \emph{partial} cost volume with a small neighborhood at each recurrence to capture large displacements sequentially.}

In this paper, we propose an alternative to these prevailing approaches, which does not use the sequential estimation strategy to process the cost volume. Instead, cost volumes with different dilation rates are constructed at the same time. Even with a small search neighborhood, both small and large displacements can be captured simultaneously. By concatenating such cost volumes together, a simple convolutional network (a U-Net) is then employed to process the dilated cost volumes only once to obtain the optical flow. Specifically, we estimate the interpolation weights between all possible displacements captured in the dilated cost volumes to get the optical flow. In addition to computing the loss for the optical flow, we also design a loss term to better supervise the interpolation weights, leading to better accuracy.

Compared with other approaches, such as PWC-Net~\cite{sun2018pwc} and RAFT~\cite{teed2020raft}, our approach is conceptually simpler, and does not require  sequential processing of the cost volumes, as shown in Fig.~\ref{fig:teaser}. While obtaining comparable error rates on standard benchmarks, our approach runs significantly faster at inference time, achieving 30 fps (frames per second) for a Sintel-resolution image (with a size of $1024\times436$) on a mid-end 1080ti GPU. Our code as well as the model weights will be made publicly available.

\section{Related Work}
In this section, we discuss previous optical flow methods. Due to space limits, we focus on neural network-based approaches.

FlowNet, proposed in~\cite{Dosovitskiy:2015Flownet}, has two variants, FlowNetS and FlowNetC, both of which have an encoder-decoder structure. 
FlowNetS simply \emph{concatenates} the feature representations of the two images obtained from the encoder and lets the decoder learn how to compute optical flow. In contrast, FlowNetC constructs a cost volume by computing matching costs (or similarity) between two feature maps. To improve the accuracy, especially for large displacements, FlowNet2~\cite{Ilg:2016:Flownet2} concatenates the FlowNetS and FlowNetC variants in a cascade, where the optical flow estimation is progressively refined. This is the first instance where a neural network reports better or on-par optical flow results with classical engineered approaches.

Although FlowNet2 achieves good accuracy, it has 162M parameters. The more compact SpyNet is proposed in~\cite{Ranjan:2016:SpyNet}. It computes flow in a coarse-to-fine manner by using a \emph{pyramid} structure borrowed from classical approaches. 
PWC-Net~\cite{sun2018pwc} extends the pyramid structure used in SpyNet. In each pyramid level, a cost volume is built by \emph{warping} the second image's feature map using the optical flow estimation in the previous level. As a result, large displacements can be captured in a sequential manner.
A similar coarse-to-fine strategy is used in other approaches. LiteFlowNet~\cite{Hui_2018_CVPR} also uses a pyramid structure to estimate optical flow in a cascade manner and proposes a flow regularization layer. In the recent extension LiteFlowNet3~\cite{hui2020liteflownet3}, an adaptive modulation prior is added to the cost volume, and local flow consistency is used to improve the final accuracy. HD$^3$~\cite{yin2019hierarchical} converts optical flow estimation into discrete distribution decomposition. SENSE~\cite{jiang2019sense} extends PWC-Net to solve optical flow and stereo disparity at the same time with a shared encoder. In~\cite{yang2019volumetric}, a separable 4D convolution is proposed to process the cost volume, which is converted into two successive 3D convolutions. In~\cite{wang2020displacement}, 2D convolutions are independently applied to each sampled displacement in the cost volume for displacement-invariant cost learning (DICL). In~\cite{xu2022gmflow}, optical flow estimation is modeled as a global matching problem by computing the similarities of features, where iterative refinement shows improved accuracy.

\new{Instead of using a feature hierarchy for coarse-to-fine estimation, an RNN is used in RAFT~\cite{teed2020raft}. It builds a full-range cost volume capturing the similarity between all pairs of locations between two images. But at each recurrence step, only a partial cost volume in a small neighborhood is used to estimate an offset. This offset is used to move the estimated optical flow (displacement) iteratively closer to the optimum.} In~\cite{jiang2021learning}, sparse matches instead of a full cost volume is used to reduce the memory consumption. In a similar effort~\cite{xu2021high}, 1D attention and correlation is used so that the RAFT can be used for high-resolution images. Zhang~\etal propose to use a separable cost volume module using non-local aggregation layers to reduce motion ambiguity~\cite{zhang2021separable}. Kernel patch attention is used to better use the local affinity to implicitly enfore the smoothness constratint~\cite{luo2022learning}. DIP~\cite{zheng2022dip} uses an new inverse propagation inspired by the classical PatchMatch algorithm to better estimate the cost volume. CRAFT proposes to replace the dot-product correlations with transformer cross-frame attention~\cite{sui2022craft}. In~\cite{zhao2022global}, argmax is applied on the 4D cost volume to efficiency compute the global matching to better capture large displacement. In~\cite{bai2022deep}, deep equilibrium (DEQ) flow estimators are proposed to replace the RNN.

Such \new{sequential estimation} approaches are inherently slow as optical flow estimation at each pyramid level or recurrence step is dependent on the results in the previous one. In contrast, 
Unlike these coarse-to-fine or recurrence-based approaches, we build cost volumes with different dilation factors to effectively capture small and large displacements simultaneously. Consequently, our approach does not need the sequential estimation strategy.

There are other approaches, whose efforts are complementary to ours. A set of improvements about model training protocols, including the data sampling process, model regularization, and data augmentation, are presented in ScopeFlow~\cite{Bar-Haim20scope}. A learnable cost volume is proposed in~\cite{xiao20learnable}, which considers the effectiveness of different feature channels by assigning different weights to different channels. Sun~\etal propose to learn to generate training data to train optical flow models~\cite{sun2021autoflow}. Depth learned using a monocular depth estimation model is used to generate optical flow with a virtual camera in~\cite{aleotti2021learning}. Detail-preserving residual feature pyramid modules are proposed in, which retains important details in the feature maps to better compute the cost volume~\cite{long2022detail}. Self-supervised consistency loss are proposed in~\cite{jeong2022imposing} to improve an optical flow model's accuracy.

\new{Using dilations in cost volumes is not completely new. FlowNetC~\cite{Dosovitskiy:2015Flownet} only uses a single dilation factor of 2, which does not fully exploit the potential of using dilations to capture large displacements. In Devon~\cite{lu2020devon}, dilated cost volumes are used as a replacement for the warping modules in a sequential coarse-to-fine estimation model. {By sharp contrast to FlowNetC~\cite{Dosovitskiy:2015Flownet}, we use multiple dilation factors to better capture small and large displacements. Additionally, unlike Devon~\cite{lu2020devon}, we use dilated cost volumes as an alternative for the sequential estimation strategy to compute optical flow. Moreover, our model achieves significantly better accuracy than both FlowNetC~\cite{Dosovitskiy:2015Flownet} and Devon~\cite{lu2020devon}.}}

\section{Dilated Cost Volume Networks}

\newcommand{\vx}{\mathbf{x}}
\newcommand{\vc}{\mathbf{c}}
\subsection{Dilated Cost Volumes}
\label{sec:d3dcv}
Given two input images $I_1$ and $I_2$ with height $H$ and width $W$, we extract their $L_2$-normed feature representations $\vx_1^s$ and $\vx_2^s$ at stride $s$ using a CNN, where $s$ corresponds to the spatial resolution downsample factor w.r.t.~the input images. To search for the correct correspondence for a position $p=(x, y)$ in $\vx_1^s$, we need to compare its feature vector with that of a candidate position $q=(x+u, y+v)$ in $\vx_2^s$, where $u$ and $v$ are the offsets of the pixel from $p$ to $q$. To measure the similarity between feature vectors at $p$ and $q$, we have
\begin{align}
    \vc^s(u, v, x, y) = f\left(\vx_1^s(x, y),\vx_2^s(x+u, y+v)\right),
\end{align}
where $f(\cdot, \cdot)$ is a function measuring the similarity between two feature vectors. Here we divide each of the vectors $\vx_1^s$ and $\vx_2^s$ into $C$ sub-vectors and compute the cosine similarity between each pair of corresponding sub-vectors. The output of $f$ therefore has $C$ dimensions.  As we need to sample in a local 2D neighborhood for all possible correspondences, we have $u\in[-k, k]$ and $v\in[-k, k]$, where $k$ is the neighborhood radius. The full cost volume size is thus $C\times U\times V\times\frac{H}{s}\times \frac{W}{s}$, where $U=V=2k+1$.

\begin{figure}
\centering
\renewcommand{\tabcolsep}{0.25mm}
\begin{tabular}{ccc}
\includegraphics[height=0.25\linewidth]{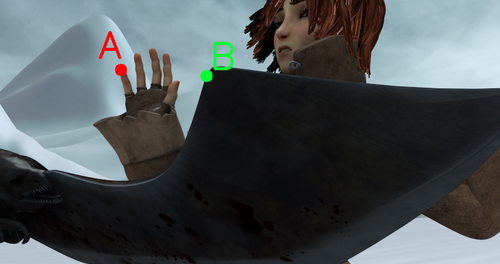} & 
\includegraphics[height=0.25\linewidth]{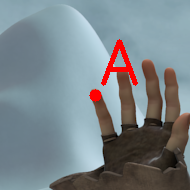} &
\includegraphics[height=0.25\linewidth]{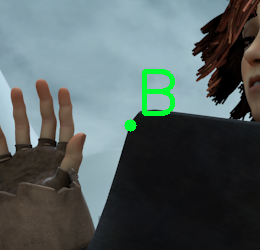} \\
\includegraphics[height=0.25\linewidth]{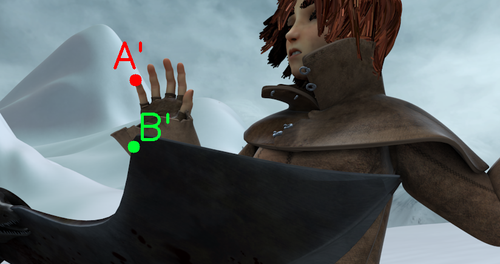} &
\includegraphics[height=0.25\linewidth]{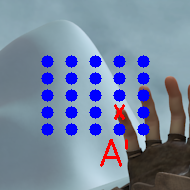} &
\includegraphics[height=0.25\linewidth]{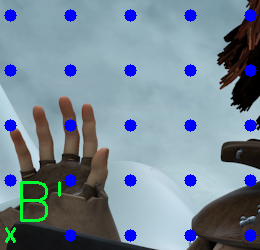} \\
(a) input images & (b) dilation=1 & (c) dilation=5 \\
\end{tabular}
\caption{\textbf{Illustration of using dilation to capture both small and large displacements.} (a) input two images where points $A$ and $B$ move to $A'$ and $B'$, respectively. (b) two patches around $A$ in two images. (c) two patches around $B$ in two images. Blue dots in (b) and (c) correspond to candidate displacements when constructing cost volumes. With a small search radius (2 in this example), correct displacements (denoted by red and blue crosses, respectively) can be captured using two different dilation factors. Best viewed in color.}
\label{fig:illustration_dilation}
\end{figure}

Due to the striding factor, such a cost volume $\vc^s$ captures candidate horizontal displacements\footnote{In this paper, we use ``displacement'' to denote a pixel's offset over two input images and ``correspondence'' to refer to offsets over two feature maps. So a displacement is the multiplication of a correspondence by the stride of the feature map.} 
across two input images in the range of $s\odot[-k, k]$, where $\odot$ denotes the elementwise multiplication between a scalar and a vector. For simplicity, we use only the horizontal displacement for illustration here (vertical displacement can be analyzed similarly). To account for large displacements, which is critical for accurate optical flow estimation, either a larger stride $s$ or neighborhood radius $k$ can be used. Both of them are problematic, however. A larger stride means more downsampling and loss of spatial resolution. A large neighborhood radius, on the other hand, results in a large cost volume and heavy computation.

\begin{figure*}[t]
\centering
\includegraphics[width=0.75\linewidth]{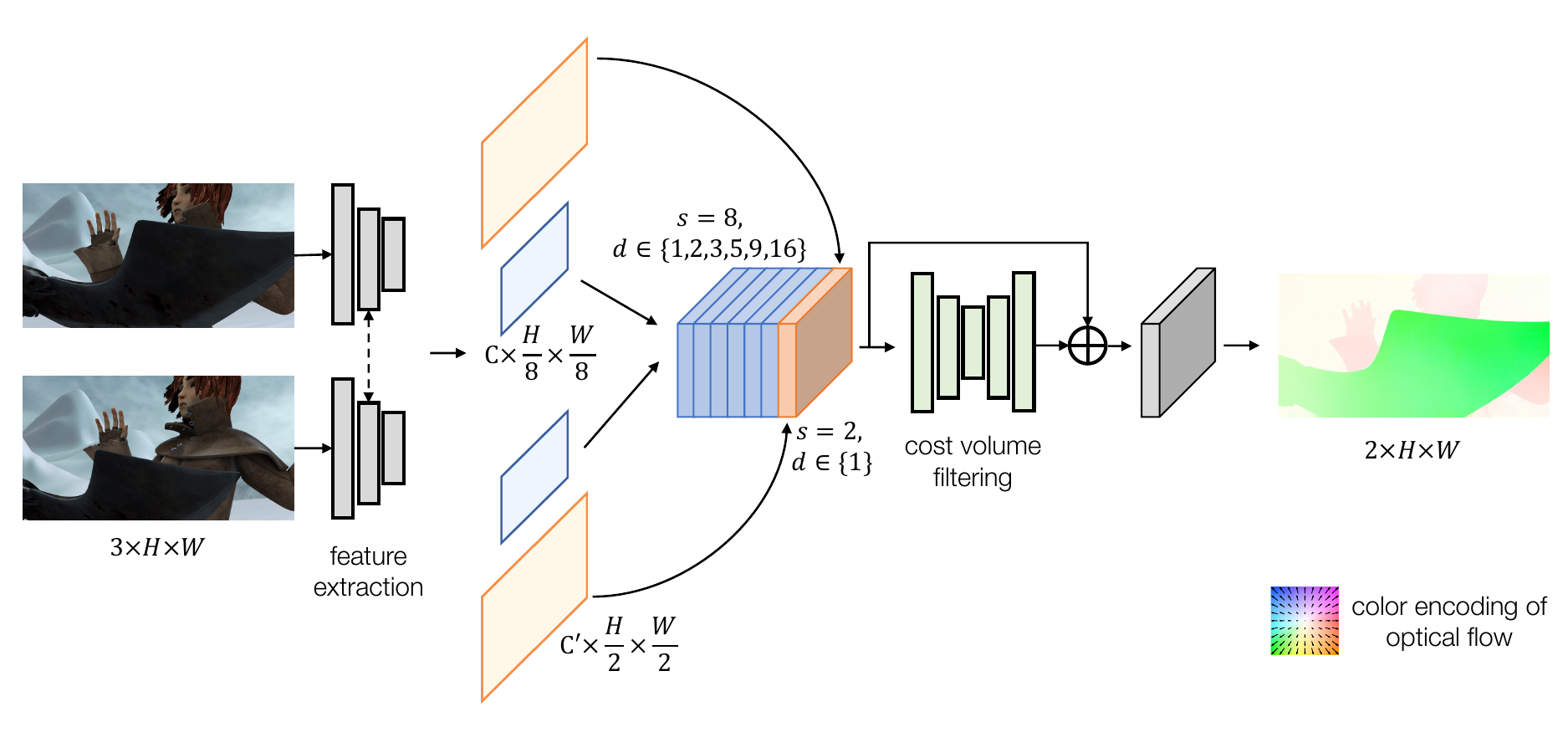}
\caption{\textbf{Pipeline of DCVNet.} Feature representations of two input images are obtained from the encoder, which are used to construct the dilated cost volumes at different strides and dilation rates. A U-Net is employed to process the cost volumes to produce a set of interpolation weights over the captured displacements in the cost volume to compute the optical flow.}
\label{fig:dcv_pipeline}
\end{figure*}

Instead, we propose to use dilation factors to construct cost volumes to deal with small and large displacement at the same time. Specifically, we have
\begin{align}
\vc^{s,d}&(u^{d}, v^{d}, x, y) = f\left(\vx_1^s(x, y),\vx_2^s(x+u^{d}, y+v^{d})\right),\nonumber\\
&\text{where}~u^{d}\in d\odot[-k, k], v^{d}\in d\odot[-k, k].
\end{align}
Here $d$ is a dilation factor. Now the search range of displacement over two input images is $sd\odot[-k, k]$. In this way, we can capture large displacements by having a large $d$ while maintaining small $k$ and $s$, which preserves both computational efficiency and spatial resolution for the cost volume. Fig.~\ref{fig:illustration_dilation} illustrates how dilation helps capture both small and large displacements with a small neighborhood radius. Specifically, in this paper, we consider $s=8$ and $d\in\{1,2, 3,5,9,16\}$. As we can see in Table~\ref{tab:candidate_placement}, a displacement as large as 512 pixels can be captured using a dilation factor $d=16$, stride $s=8$, and neighborhood radius $k=4$.

As the dilation factors increase, the gap between candidate displacements also gets larger. To deal with this issue, we also build a cost volume with $s=2$ and $d=1$ to capture small and fine displacement. We do spatial sampling of 4 to make the spatial resolution compatible with the cost volumes constructed over the stride of 8. Finally, we concatenate all cost volumes over different strides and dilation factors. Our final cost volume has a dimension of $C'\times \frac{H}{8}\times\frac{W}{8}$, where $C'=D\times C\times U\times V$ and $D$ is the total number of dilation factors ($D=7$ in our case). 

\begin{table}[t]
\small
\caption{\textbf{Displacements over input images} captured using different strides and dilation factors.}
\label{tab:candidate_placement}
\renewcommand{\tabcolsep}{2pt}
\begin{tabular}{c|c|c}
\toprule
stride & dilation & \multirow{2}{*}{candidate horizontal displacements} \\
($s$) & ($d$) & \\
\midrule
2 & 1 & \{-8, -6, -4, -2, 0, 2, 4, 6, 8\} \\
\midrule
8 & 1 & \{-32, -24, -16, 8, 0, 8, 16, 24, 32\} \\
8 & 5 & \{-160, -120,  -80,  -40,    0,   40,   80,  120,  160\}\\
8 & 9 & \{-288, -216, -144,  -72,    0,   72,  144,  216,  288\}\\
8 & 16 & \{-512, -384, -256, -128, 0, 128, 256, 384, 512\} \\
\bottomrule
\end{tabular}
\end{table}

\begin{figure}
\centering
\renewcommand{\tabcolsep}{0.28mm}
\begin{tabular}{cc}
    \begin{tabular}{c}
        \includegraphics[height=0.25\linewidth]{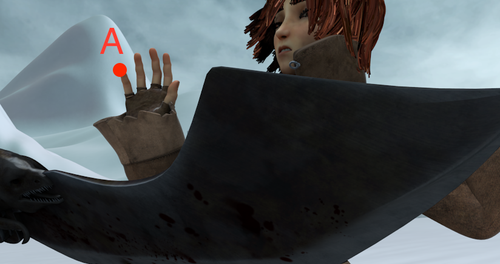} \\
        ~ \\
        ~\\
        \includegraphics[height=0.25\linewidth]{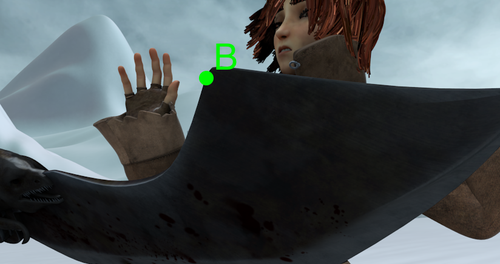} \\
        ~
    \end{tabular}
    &
    \begin{tabular}{cccc}
        \includegraphics[height=0.12\linewidth]{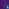} &
        \includegraphics[height=0.12\linewidth]{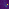} &
        \includegraphics[height=0.12\linewidth]{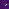} &
        \includegraphics[height=0.12\linewidth]{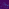} \\
        \includegraphics[height=0.12\linewidth]{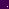} &
        \includegraphics[height=0.12\linewidth]{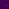} &
        \includegraphics[height=0.12\linewidth]{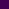} &
        \includegraphics[height=0.12\linewidth]{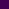} \\
        $d=1$ & $d=2$ & $d=3$ & $d=5$ \\
        ~ & ~ & ~ & ~ \\
        \includegraphics[height=0.12\linewidth]{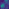} &
        \includegraphics[height=0.12\linewidth]{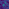} &
        \includegraphics[height=0.12\linewidth]{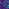} &
        \includegraphics[height=0.12\linewidth]{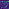} \\
        \includegraphics[height=0.12\linewidth]{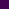} &
        \includegraphics[height=0.12\linewidth]{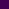} &
        \includegraphics[height=0.12\linewidth]{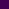} &
        \includegraphics[height=0.12\linewidth]{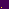} \\
        $d=1$ & $d=2$ & $d=3$ & $d=5$ \\
    \end{tabular} \\
\end{tabular}
\caption{\textbf{Illustration of interpolation weights.} For both points A and B, in the right, we show the interpolation weights obtained with and without the U-Net filtering on the top and bottom, respectively. Each image represents $U\times V$ ($9\times 9)$ interpolation weights. The feature stride is 8 and different dilation factors are shown in the bottom. We can see that for the point A, whose motion magnitude is small, a small dilation factor is sufficient to capture the correspondence. While for the point B, whose motion magnitude is large, a large dilation factor is needed. (Color encoding: blue is close to 0 and yellow is close to 1. Best viewed in color.)}
\label{fig:interpolation_weights}
\end{figure}

\subsection{Cost Volume Filtering for Optical Flow}
\label{sec:cv_to_flow}
So far, we have introduced our dilated cost volumes.
How can we translate such cost volumes into exact pixel-wise displacements, \ie, optical flow? Instead of directly \emph{regressing} optical flow values, we do \emph{interpolation} between all possible displacements similar to~\cite{yang2019volumetric,wang2020displacement}. Specifically, we have
\newcommand{\vf}{\mathbf{f}}
\newcommand{\vu}{\mathbf{u}}
\begin{align}
\vf=\sum_{i,s,d} \omega_{i,s,d} \vf_{i,s,d},
\label{eq:flow_hypothesis}
\end{align}
where $\sum_{i,s,d}\omega_{i,s,d}=1$ and $\omega_{i,s,d}\geq0$. 
$\vf_{i,s,d}=\left(\mu_{i,s,d}, \nu_{i,s,d}\right)$ is a single 2D displacement at stride $s$ with dilation $d$, where $\mu_{i,s,d}\in sd\odot[-k, k], \nu_{i,s,d}\in sd\odot[-k, k]$. At a particular stride, for each dilation factor, there are $UV$ such sampled displacements in a cost volume. \new{To obtain the interpolation weights, we use a U-Net taking our dilated cost volumes as input to estimate $\omega_{i,s,d}$, where a skip connection from the cost volume to the output is added.} A \texttt{softmax} is performed on top of the U-Net's output to ensure that the constraints on $\omega_{i,s,d}$ are satisfied.

The architecture of our proposed dilated cost volume network (DCVNet) is illustrated in Fig.~\ref{fig:dcv_pipeline}. We use a feature encoder similar to that used in~\cite{teed2020raft}, except we only use a single residual block in the stride of 2, with Instance Normalization~\cite{ulyanov2016instance} layers to extract features of input images to construct the cost volume. We empirically found that having another context encoder is not significantly helpful and yet substantially increases the number of parameters. There are no normalization layers for the rest of the DCVNet. We use Leaky ReLUs with a slope of 0.1 for the entire network. We use the same convex upsampling strategy used in~\cite{teed2020raft} to upsample estimated optical flow to the input's resolution. We provide more details of the network architecture in the supplementary material.

\subsection{Loss Function}
Denote the estimated optical flow before and after the upsampling as $\hat{\vf}'$ and $\hat{\vf}$, respectively, and the ground-truth as $\vf$. We use the \texttt{L1} loss to supervise the network training.
\newcommand{\calL}{\mathcal{L}}
\begin{align}
    \calL_f = \alpha||\hat{\vf}' - \vf'||_1 + ||\hat{\vf} - \vf||_1,
\end{align}
where $\vf'$ is the downsampled ground truth of $\vf$, which has the same resolution as $\hat{\vf}'$. $\alpha$ is empirically set as 0.25.

At the same time, we found that adding extra constraints to the interpolation weights $\omega_{i,s,d}$ leads to better results. Note there are many plausible solutions of $\omega_{i,s,d}$ that yield the same optical flow. To add constraints of the interpolation weights for each pixel, we compute the bilinear interpolation weights over the four nearest displacement vectors surrounding the ground-truth optical flow value. We use the \texttt{CrossEntropy} loss between the estimated $\hat{\omega}_{i,s,d}$ and the generated ground-truth $\omega_{i,s,d}$.
\begin{align}
    \calL_{\omega} = -\sum_{i,s,d}\omega_{i,s,d}\log\hat{\omega}_{i,s,d}.
\end{align}

The final loss is defined as
\begin{align}
    \calL = \calL_f + \beta\calL_{\omega},
\end{align}
where $\beta$ balances $\calL_f$ and $\calL_{\omega}$. We empirically found that it leads to better accuracy by annealing $\beta$ using a cosine schedule, where the initial value is 1 and the final value is 0. We hypothesize that at the beginning of the training, adding the strong prior using the bilinear interpolation weights to $\omega_{i,s,d}$ helps the training but it becomes less effective as  training goes on.

\section{Experiments}

\subsection{Implementation Details}
\noindent\textbf{Pre-training.} We train our model on the synthetic SceneFlow dataset~\cite{mayer2016a} following~\cite{jiang2019sense}, which consists of FlyingThings3D, Driving, and Monkaa. We found using FlyingChairs~\cite{Dosovitskiy:2015Flownet} and FlyingThings leads to worse results for our model. Only optical flow annotations are used for training. Interestingly, such a pre-training results in worse results for RAFT (3.16 vs 2.71 in terms of average end-point-error on the final pass of the MPI-Sintel training set~\cite{Butler:ECCV:2012}).

\newcommand{\bd}[1]{\textbf{#1}}
\begin{table}[t]
\caption{\textbf{Average EPE results on MPI Sintel optical flow dataset.}  ``-ft'' means fine-tuning on the MPI Sintel \emph{training} set. The numbers in parentheses are results on the data the methods have been fine-tuned on. They are not directly comparable and put here for completeness. $^\dagger$ indicates a model uses extra training data. 
} 	
\label{tab:flow_sintel}
\footnotesize
\centering
\setlength\tabcolsep{4.5pt}
\begin{tabular}{l|cc|cc|cc}
    \toprule
	\multirow{2}{*}{Methods} & \multicolumn{2}{c|}{Training}  & \multicolumn{2}{c|}{Test} & Time & \#Para\\
	\cmidrule{2-3} \cmidrule{4-5} \cmidrule{6-7} 
	& Clean &  Final & Clean &  Final &  (s) & (M) \\
	\midrule
	FlowNet2~\cite{Ilg:2016:Flownet2} & 2.02 &   3.14	 &   3.96 & 6.02 &0.12 & 162\\
	PWC-Net~\cite{sun2018pwc} & 2.55 &	3.93 & - & - & 0.04 & 8.8 \\
	LiteFlowNet~\cite{Hui_2018_CVPR}  & 2.48 & 4.04 & - & - &0.07 & 5.4 \\
	LiteFlowNet2 & 2.24 & 3.78 & - & - & \textbf{0.03} & - \\
	FlowNet3~\cite{Ilg2018occlusions} &  2.08 & 3.94 & 3.61 & 6.03 & 0.07 & 117 \\
	HD$^3$~\cite{yin2019hierarchical} & 3.84 & 8.77 & - & - & 0.14 & 38.6\\
	SENSE~\cite{jiang2019sense} & 1.91 & 3.78 & - & - & 0.04 & 8.3 \\
	VCN~\cite{yang2019volumetric} & 2.21 & 3.68 & - & - & 0.26 & 6.2 \\
	MaskFlow~\cite{zhao2020maskflownet} & 2.25 & 3.61 & - & - & - & -\\
    Devon~\cite{lu2020devon} & 2.45 & 3.72 & - & - & 0.04 & - \\
    DICL~\cite{wang2020displacement} & 1.94 & 3.77 & - & - & 0.08 & 9.8 \\
    RAFT-small~\cite{teed2020raft} & 2.21 & 3.35 & - & - & 0.05 & \bf{1.0} \\
	RAFT~\cite{teed2020raft} &  \bd{1.43} & \bd{2.71} & - & - & 0.3 & 5.3 \\
	Ours & 1.91 & 3.28 & - & - & \bd{0.03} & 7.9\\
	\midrule
	FlowNetS-ft~\cite{Dosovitskiy:2015Flownet} & (3.66)  & (4.44)& 6.96  & 7.52 & 0.02 & 38.7 \\
	FlowNetC-ft~\cite{Dosovitskiy:2015Flownet}  & (3.50) & (3.89) & 6.85 & 8.51 & 0.03 & 39.1  \\
	SpyNet-ft~\cite{Ranjan:2016:SpyNet} & (3.17) & (4.32) & 6.64 & 8.36  &0.16 & 1.2 \\
	FlowNet2-ft~\cite{Ilg:2016:Flownet2} & ({1.45})	&  ({2.01}) &  4.16 & 5.74 &0.12 & 162	\\ %
	PWC-Net-ft~\cite{sun2018pwc} & ({1.70})	& ({2.21}) & {3.86}  & 5.13    & 0.04 & 8.8  \\	
	LiteFlowNet-ft~\cite{Hui_2018_CVPR} & (1.45) & (1.78) & 4.54 & 5.38 & 0.07 & 5.4 \\
	LiteFlowNet2-ft~\cite{hui2018arxiv} & (1.30) & (1.62) & 3.48 & 4.69 & 0.03 & - \\
	LiteFlowNet3-ft~\cite{hui2020liteflownet3} & (1.32) & (1.76) & 2.99 & 4.45 & 0.05 & 5.2 \\
	FlowNet3-ft~\cite{Ilg2018occlusions} &  (1.47) & (2.12) & 4.35 & 5.67 & 0.07 & 117 \\
	HD$^3$-ft~\cite{yin2019hierarchical} & (1.87) & (1.17) & 4.79 & 4.67 & 0.14 & 38.6 \\
	SENSE-ft~\cite{jiang2019sense} & (1.54) & (2.05) & 3.60 & 4.86 & 0.04 & 8.3 \\
	VCN-ft$^\dagger$~\cite{yang2019volumetric} & (1.66) & (2.24) & 2.81 & 4.40 & 0.26 & 6.2 \\
	MaskFlow-ft$^\dagger$~\cite{zhao2020maskflownet} & - & - & 2.52 & 4.17 & - & - \\
	Devon-ft~\cite{lu2020devon} & (1.97) & (2.67) & 4.34 & 6.35 & 0.04 & - \\
	DICL-ft~\cite{wang2020displacement} & (1.11) & (1.60) & 2.12 & 3.44 & 0.08 & 9.8 \\
	RAFT-ft$^\dagger$~\cite{teed2020raft} & (0.77) & (1.27) & \bd{1.61} & \bd{2.86} & 0.3 & 5.3 \\
	Ours-ft$^\dagger$ & (1.04) & (1.37) & 2.36 & 3.66 & \bd{0.03} & 7.9 \\ 
	\bottomrule
\end{tabular}
\end{table}

During training, we closely follow the setting used in RAFT~\cite{teed2020raft}. Specifically, we use extensive data augmentations including color jittering, random crops, random resizing, and random horizontal and vertical flips. The crop size is $400\times 720$. The DCVNet is trained for 800K iterations with a batch size of 8 using the AdamW optimizer~\cite{kingma2014adam}. The initial learning rate is 0.0002 and is updated following the OneCycle learning rate schedule~\cite{smith2017super} with a linear annealing strategy and a warmup factor of 0.05. We also perform gradient norm clip with a value of 1. 

\begin{figure*}[t]
\centering
\renewcommand{\tabcolsep}{0.8pt}
\newcommand{\loadFig}[1]{\includegraphics[width=0.19\linewidth]{#1}}
\begin{tabular}{ccccc}
\loadFig{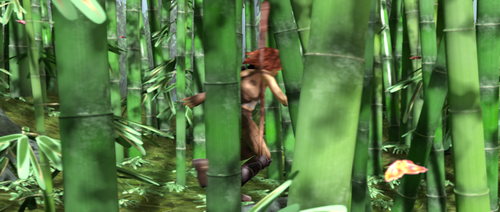} &
\loadFig{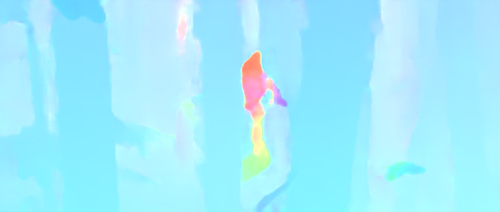} &
\loadFig{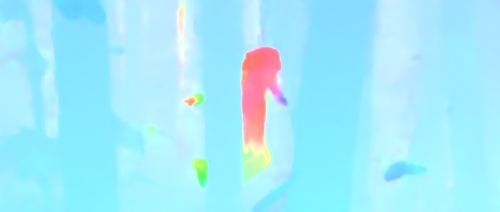} &
\loadFig{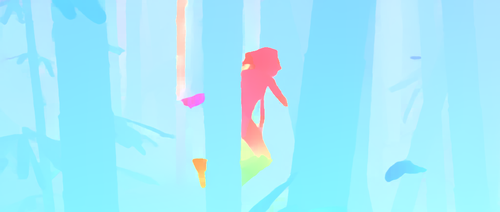} &
\loadFig{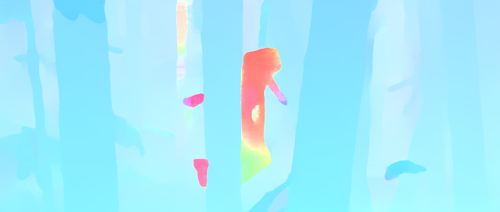} \\
\loadFig{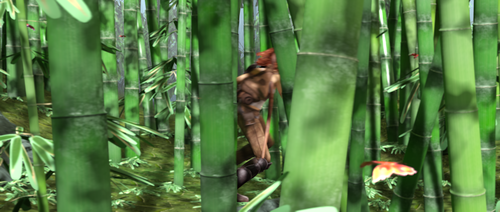} &
\loadFig{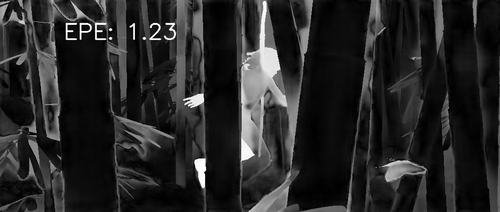} &
\loadFig{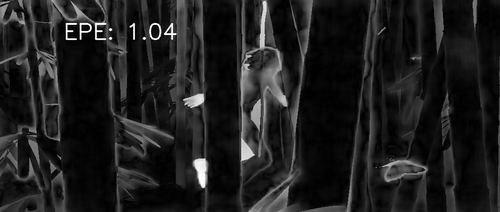} &
\loadFig{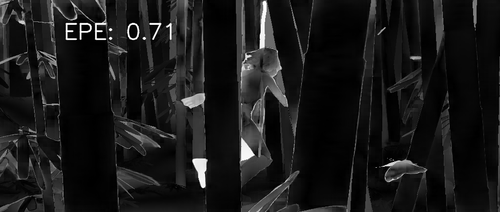} &
\loadFig{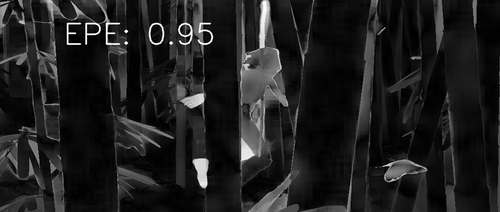} \\
\loadFig{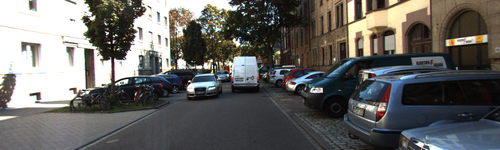} & 
\loadFig{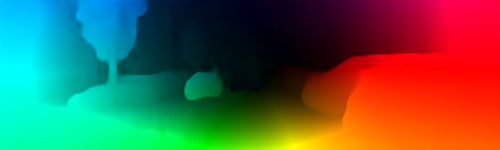} & 
\loadFig{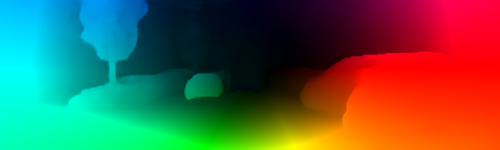} & 
\loadFig{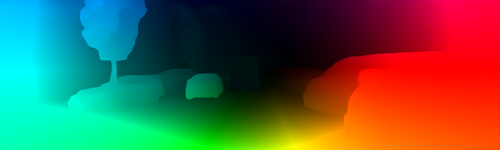} & 
\loadFig{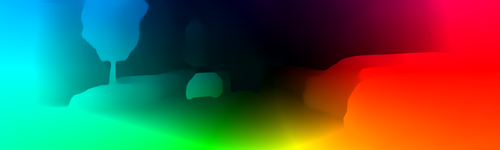} \\ 
\loadFig{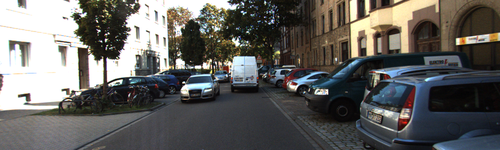} & 
\loadFig{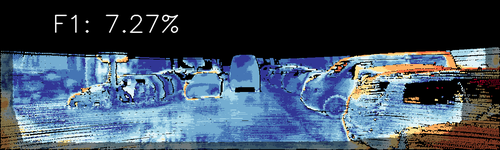} & 
\loadFig{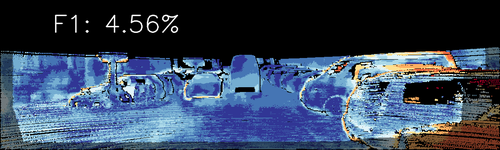} & 
\loadFig{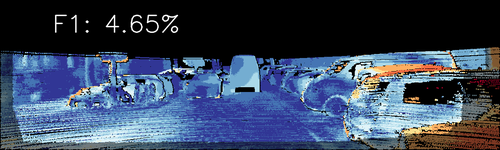} & 
\loadFig{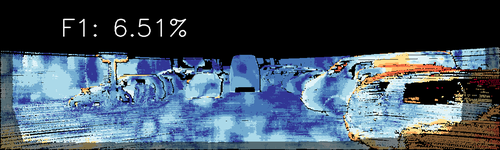} \\ 
(a) input images & (b) PWCNet~\cite{sun2018pwc} & (c) VCN~\cite{yang2019volumetric} & (d) RAFT~\cite{teed2020raft} & (e) our DCVNet \\
\end{tabular}
\caption{\textbf{Visual comparison of optical flow estimations}. From left to right: (a) input images, (b) PWCNet~\cite{sun2018pwc}, (c) VCN~\cite{yang2019volumetric}, (d) RAFT~\cite{teed2020raft}, and (e) our DCVNet. For each method, we show colorized optical flow and error maps (obtained from online servers). For the error maps, white and red indicate large error while black and blue mean small error. Best viewed in color.}
\label{fig:vis_flow}
\end{figure*}

\begin{table}[t]
\caption{\textbf{Results on the KITTI optical flow dataset.} ``-ft'' means fine-tuning on the KITTI \emph{training} set and the numbers in the parenthesis are results on the data the methods have been fine-tuned on.} 	
\label{tab:flow_kitti}
\footnotesize
\centering
\setlength\tabcolsep{2pt} 
\begin{tabular}{l|ccc|ccc}
    \toprule
	\multirow{3}*{Methods} & \multicolumn{3}{c|}{ KITTI 2012} & \multicolumn{3}{c}{ KITTI 2015}  \\
	\cmidrule{2-4} \cmidrule{5-7}
	& AEPE & AEPE & Fl-Noc &  AEPE   & Fl-all & Fl-all \\
	& \emph{train} & \emph{test} & \emph{test}   & \emph{train} & \emph{train}  & \emph{test} \\
	\midrule
	FlowNet2~\cite{Ilg:2016:Flownet2}  &4.09& - & - & 10.06 & 30.37\% & - \\ %
	PWC-Net~\cite{sun2018pwc} & 4.14 & - &- & 10.35 & 33.67\% &-\\ %
	FlowNet3~\cite{Ilg2018occlusions} & 3.69 & - & - & 9.33 & - & - \\	
	HD$^3$~\cite{yin2019hierarchical} & 4.65 & - & - & 13.17 & 24.9\% & - \\
	SENSE~\cite{jiang2019sense} & 2.55 & - & - & 6.23 & 23.29\% & - \\
	VCN~\cite{yang2019volumetric} & - & - & - & 8.36 & 25.1\% & -  \\
	MaskFlow~\cite{zhao2020maskflownet} & - & - & - & - & 23.1\% & - \\
	Devon~\cite{lu2020devon} & 4.73 & - & - & 10.65 & - & - \\
	DICL~\cite{wang2020displacement} & - & - & - & 8.70 & 23.60\% & - \\
	RAFT-small~\cite{teed2020raft} & - & - & - & 7.51 & 26.91\%  & - \\
	RAFT~\cite{teed2020raft} & - & - & - & 5.04 & \bd{17.40\%} & - \\
	Ours & \bf{2.56} & - & - & \textbf{4.83} & 23.68\% & - \\
	\midrule
	SpyNet-ft~\cite{Ranjan:2016:SpyNet} & (4.13)  & 4.7 &12.31\% & - & - & 35.07\%  \\ %
	FlowNet2-ft~\cite{Ilg:2016:Flownet2}  & ({1.28}) & {1.8} & 4.82\% & ({2.30}) & ({8.61}\%) & {10.41} \% \\ %
	PWC-Net-ft~\cite{sun2018pwc} & ({1.45})	&  {1.7}	 &4.22\%	& ({2.16})	& ({9.80}\%) & {9.60}\%  		\\  
	LiteFlowNet-ft~\cite{Hui_2018_CVPR}  & ({1.26}) & {1.7} & - & ({2.16}) & ({8.16}\%) & {10.24} \% \\ 
	LiteFlowNet3-ft~\cite{Hui_2018_CVPR}  & ({0.91}) & \bd{1.3} & 2.51\% & ({1.26}) & ({3.82}\%) & {7.34} \% \\ 
    FlowNet3-ft~\cite{Ilg2018occlusions} & (1.19) & - & 3.45\% & (1.79) & - & 8.60\% \\	
    HD$^3$-ft~\cite{yin2019hierarchical} & (0.81) & {1.4} & \bd{2.26\%} & (1.31) & (4.10\%) & 6.55\% \\
	SENSE-ft~\cite{jiang2019sense} & (1.18) & 1.5 & 3.03\% & (2.05) & (9.69\%) & 8.16\% \\ 
	VCN-ft~\cite{yang2019volumetric} & - & - & - & (1.16) & (4.10\%) & 6.30\% \\
	MaskFlow-ft~\cite{zhao2020maskflownet} & - & - & - & - & - & 6.10\% \\
	Devon-ft~\cite{lu2020devon} & (1.29) & 2.6 & - & (2.00) & - & 14.31\% \\
	DICL-ft~\cite{wang2020displacement} & - & - & - & (1.02) & (3.60\%) & 6.31\% \\
	RAFT-ft~\cite{teed2020raft} & - & - & - & (0.63) & (1.50\%) & \bd{5.10\%} \\
	Ours-ft & (0.94) & 1.6 & 5.33\% & (1.22) & (4.41\%) & 9.62\%\\
	\bottomrule
\end{tabular}
\end{table}

\noindent\textbf{Fine-tuning.} For Sintel, we fine-tune the pre-trained model using both \emph{final} and \emph{clean} passes. Following previous works, we optionally use extra data from KITTI2015~\cite{Geiger:2012:We} and HD1K~\cite{hd1k} for training. The model is trained for 400K with a batch size of 8. The initial learning rate is set to be 0.000125 and updated following the same OneCycle scheduler as we use in the pre-training stage. For KITTI, we train the model for 400K iterations with a batch size of 8. The initial learning rate is 0.0001 and the OneCyle learning rate scheduler is used. For both Sintel and KITTI, we perform similar data augmentations used in the pre-training stage. The crop sizes for Sintel and KITTI are $368\times768$ and $336\times 944$, respectively. 

\subsection{Main Results}
\noindent\textbf{Optical flow results.} Quantitative results on the MPI Sintel and KITTI benchmark datasets of different neural network-based approaches are summarized in Table~\ref{tab:flow_sintel} and Table~\ref{tab:flow_kitti}, respectively. We can see that our approach compares favorably to other approaches before and after fine-tuning. Specifically, on the more photo-realistic final pass with factors such as lighting condition changes, shadow effects, motion blur, etc, our proposed model, DCVNet, achieves on-par end-point-error (EPE) with state-of-the-art approaches like DICL~\cite{wang2020displacement} and RAFT~\cite{teed2020raft}.
Particularly, our model outperforms Devon~\cite{lu2020devon}, which uses dilated cost volumes as a replacement of the warping module in a sequential cost-to-fine optical flow model.

\begin{table}[t]
\centering
\small
\setlength\tabcolsep{5pt} 
\caption{\textbf{Number of parameters, GPU memory consumption, and inference speed} of different optical flow models, measured on Sintel.}
\label{tab:epe_different_mag}
\begin{tabular}{c|c|c|c|c|c}
\toprule
 & PWC-Net & VCN & DICL & RAFT & Ours \\
\midrule
\#Para. (M) & 9.37 & 6.23 & 9.78 & \textbf{5.26} & 7.87\\
\#Mem. (GB) & \textbf{1.11} & 2.33 & 2.78 & 1.37 & 1.16 \\
\midrule
Speed (fps) & 25 & 3.8 & 12.5 & 3.3 & \textbf{30} \\
\bottomrule
\end{tabular}
\label{tab:num_param_mem}
\end{table}

We show some visual results of estimated optical flow from different approaches in Fig.~\ref{fig:vis_flow}. We can see that our approach DCVNet can capture the motion of challenging scenes, leading to visually appealing results akin to others. Particularly, for the bamboo images, our approach produces sharper motion boundaries and smoother motion estimates in the background  compared to both PWCNet~\cite{sun2018pwc} and VCN~\cite{yang2019volumetric}. We refer readers to the supplementary materials for more visual results.

\noindent\textbf{Model size and memory.} 
Our model achieves reasonable model compactness and memory consumption, as shown in Table~\ref{tab:num_param_mem}, compared to other state-of-the-art approaches. For GPU memory, our DCVNet needs 1.16GB, which is lower than RAFT, DICL, and VCN. 

\noindent\textbf{Inference speed.} Compared to existing approaches, 
our DCVNet runs significantly faster, meeting the real-time inference requirement. On a mid-end 1080ti GPU, our approach takes only 33ms to process two RGB images from the Sintel dataset (with a resolution of $1024\times 436$), running at 30 fps.
We use a CUDA implementation for the cost volume construction, which takes 10ms. Most of the time is spent on the feature extraction part that takes 14ms. The decoder part to convert the cost volume to optical flow needs 9ms.

\subsection{Ablation Studies}
To validate the effectiveness of dilated cost volumes and the loss term for the interpolation weights, we perform ablation studies. We train the models on the SceneFlow dataset.

\begin{figure}
\centering
\includegraphics[width=0.85\linewidth]{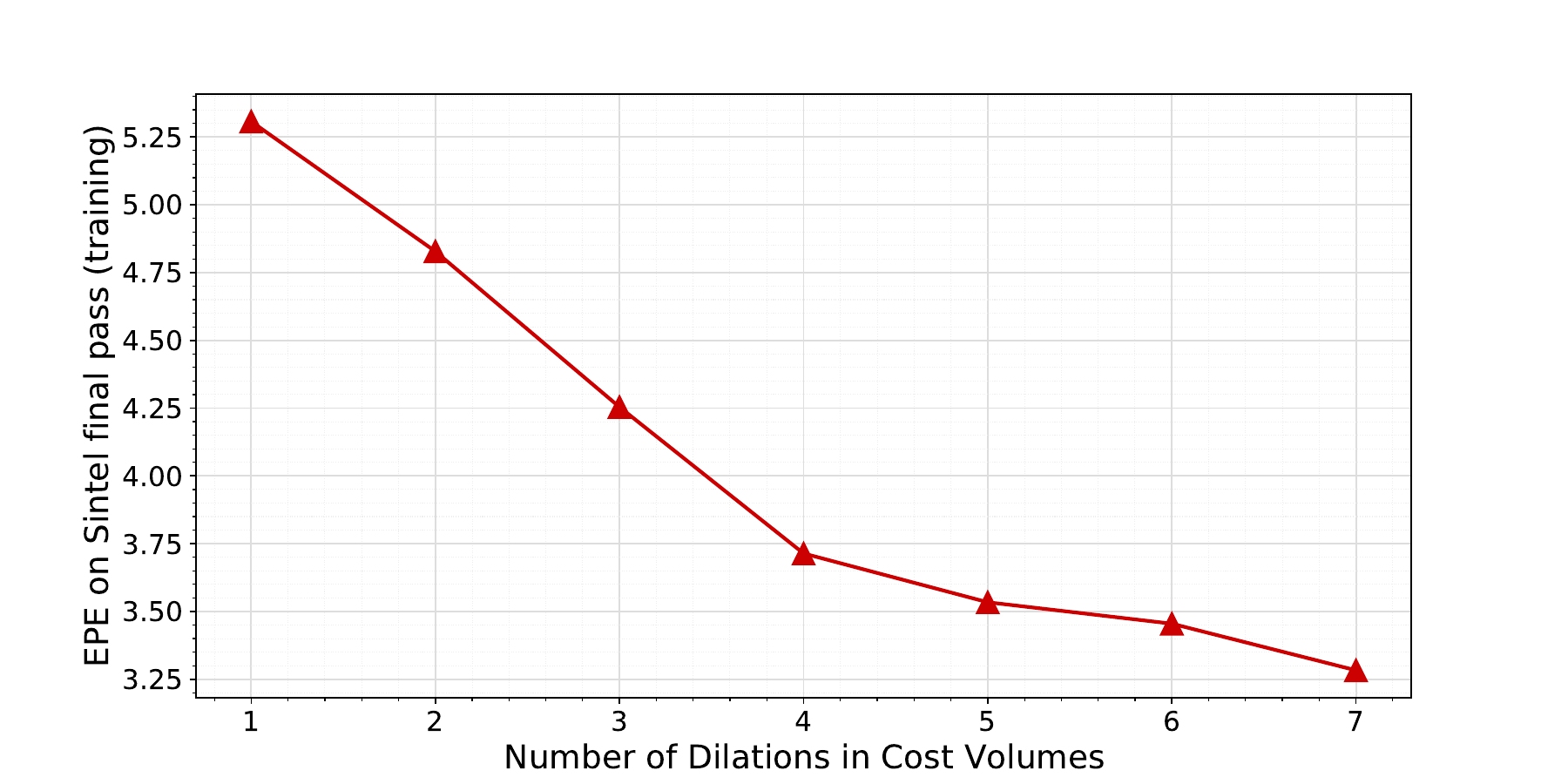}
\includegraphics[width=0.85\linewidth]{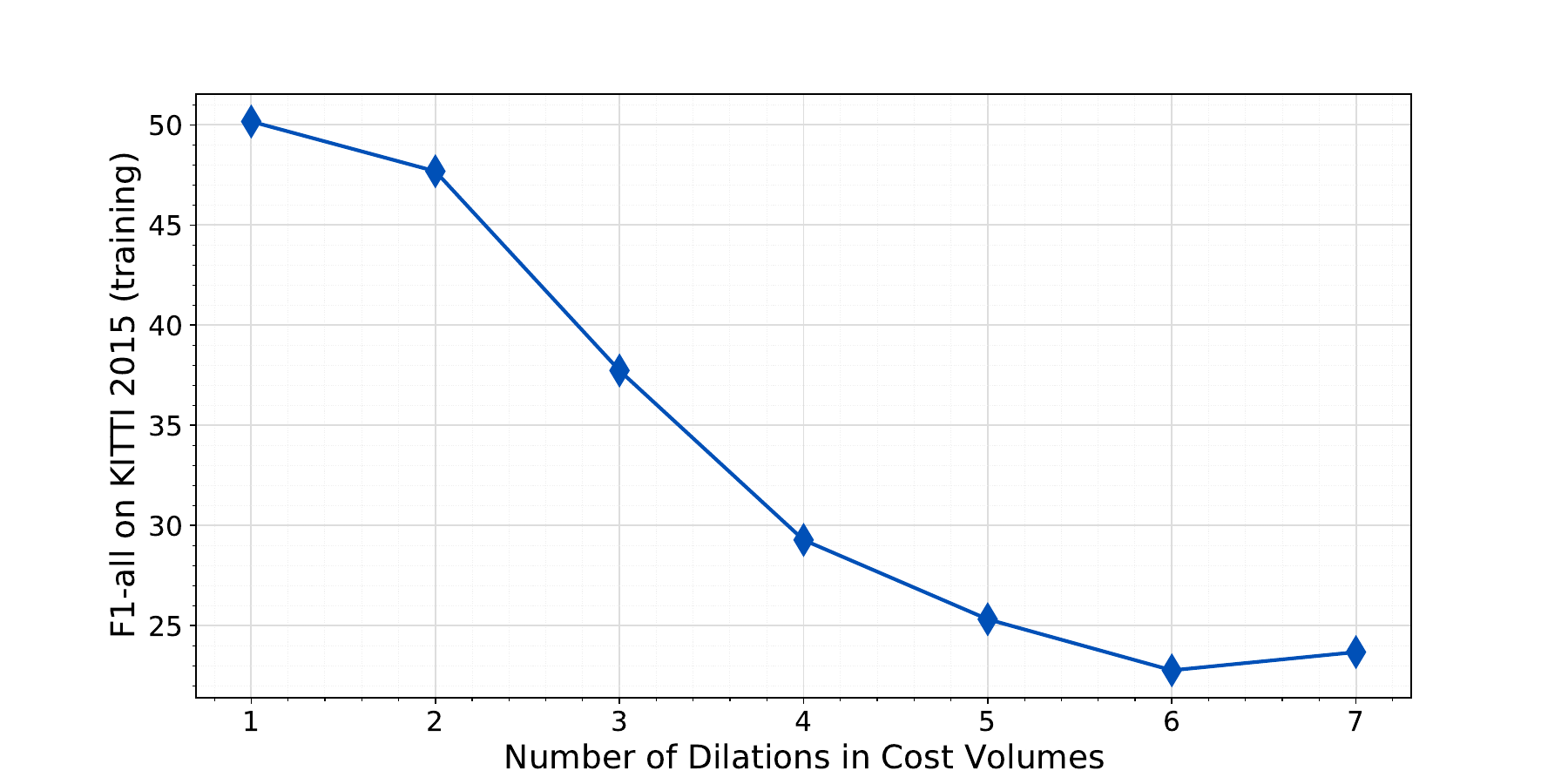}
\caption{\textbf{Effectiveness of dilation for cost volumes.} Top: EPE vs. number of dilation rates on the \emph{final} pass of the MPI Sintel training set. Bottom: F1-all error rate vs. number of dilation rates on the KITTI2015 training set.}
\label{fig:ablations_num_dilations}
\end{figure}

\noindent\textbf{Effectiveness of dilation.} We first investigate the effectiveness of using dilation for the cost volume. We vary the number of dilation rates from 7 to 1. To maintain the capability of capturing large displacement, we keep the largest stride and dilation rate and gradually remove smaller ones.  We report error rate vs. number of dilation rates in Fig.~\ref{fig:ablations_num_dilations}. We can clearly see that the error rates steadily decrease on both MPI Sintel and KITTI 2015 datasets as number of dilation increases. It validates our core idea of constructing cost volumes with dilations to deal with both small and large displacements simultaneously.

\noindent\textbf{Supervision of the interpolation weights.} We investigate the effectiveness of the loss term $\calL_{\omega}$. By setting $\beta=0$, we completely remove the supervision of the interpolation weights. On the other hand, by setting $\beta=1$, instead of annealed version, we impose strong constraints to the interpolation weights. We can see from Table~\ref{tab:ablation_loss_interp_wts}, none of them works better than the annealed $\beta$.

\section{Conclusion}
In this paper, we presented DCVNet, a dilated cost volume network for optical flow. Our core idea is to use different dilation rates to construct cost volumes to capture both small and large displacements at the same time with a small neighborhood to retain model efficiency. By doing so, our approach no longer relies on the sequential estimation strategy for optical flow, leading to a fast optical flow model. 
Our approach runs at 30fps on a mid-end 1080ti GPU and achieves comparable accuracy to existing models on standard benchmarks.

\begin{table}[t]
\centering
\small
\setlength\tabcolsep{2.55pt} 
\caption{\textbf{Effectiveness of the loss term $\calL_{\omega}$} to supervise the interpolation weights.}
\begin{tabular}{c|c|c|c|c}
\toprule
 & Sintel-clean & Sintel-final & KTTI 2012 & KITTI 2015 \\
\midrule
$\beta$=0 & 1.99 & 3.47 & 2.65 & 23.91\% \\
$\beta$=1 & \textbf{1.91} & 3.32 & 2.61 & 24.13\% \\
\midrule
annealed $\beta$ & \textbf{1.91} & \textbf{3.28} & \textbf{2.56} & \textbf{23.68\%}	\\
\bottomrule
\end{tabular}
\label{tab:ablation_loss_interp_wts}
\end{table}

{\small
\bibliographystyle{ieee_fullname}
\bibliography{egbib}
}

\newpage
\normalsize
\appendixtitleon
\appendixtitletocon
\begin{appendices}

\section{Network Architecture of DCVNet}
The overall architecture of our proposed DCVNet is illustrated in Fig.~\ref{fig:dcv_pipeline}. It consists of two major components: feature extraction for input images and cost volume filtering with a U-Net. We will introduce each of them in this section.

\begin{figure}[t]
    \centering
    \includegraphics[width=0.92\linewidth]{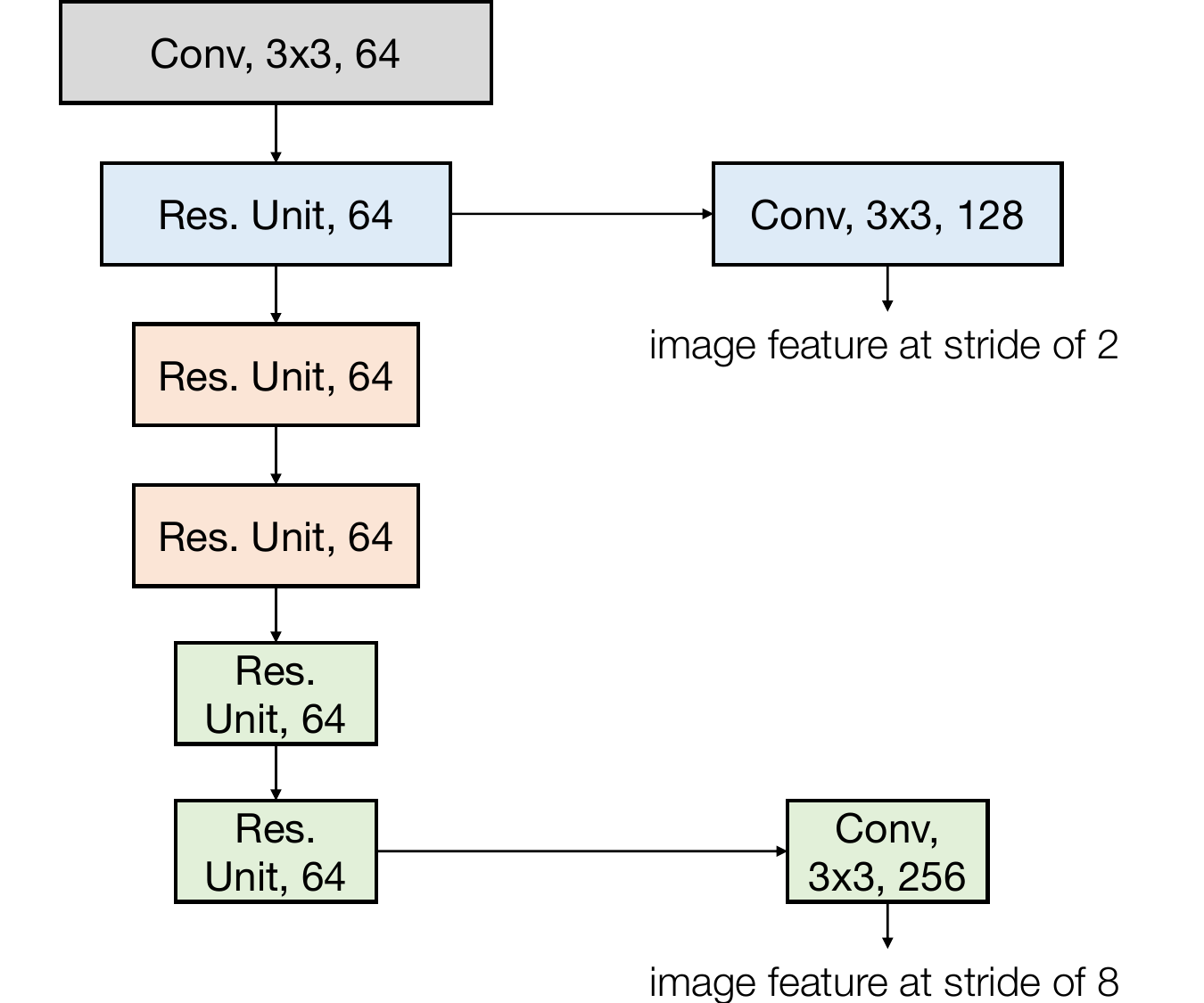}
    \caption{\textbf{Network architecture of the image encoder.} There are two output at stride of 2 and 8 with dimensions of 128 and 256, respectively. The last number indicates the number of output channels. The numbers in-between (if any) are kernel sizes.}
    \label{fig:feat_encoder}
\end{figure}

\subsection{Image Encoder}
The architecture of the image encoder is shown in Fig.~\ref{fig:feat_encoder}. We use a single encoder similar to the lightweight ResNet architecture~\cite{he16deep} used in~\cite{teed2020raft}, except we only use a single resnet block in the stride of 2. The image encoder uses Instance Normalization~\cite{ulyanov2016instance} layers, which are more effective at capturing instance-level correspondences to build the dilated cost volumes. There are two output for the image encoder at stride of 2 and 8 with dimensions of 128 and 256, respectively. 

\begin{figure}[t]
    \centering
    \includegraphics[width=0.9\linewidth]{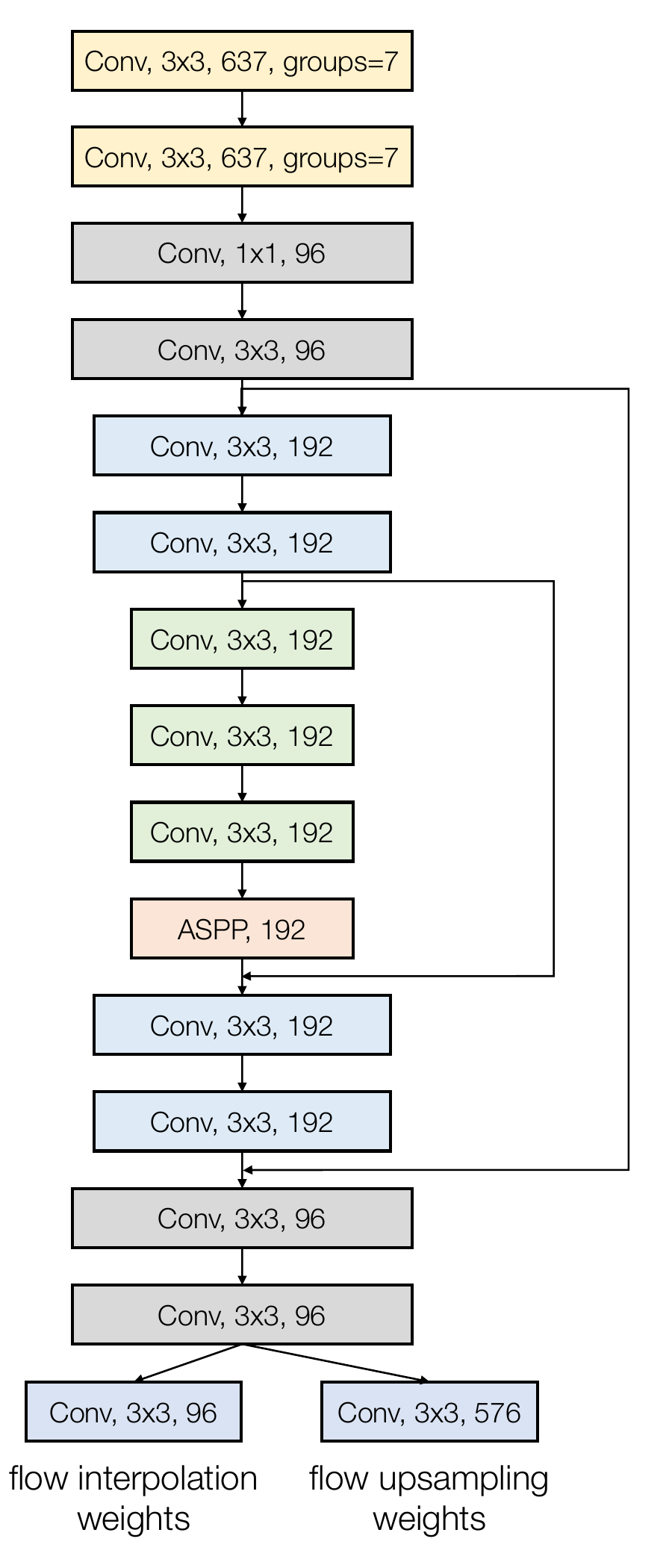}
    \caption{\textbf{Network architecture of the U-Net for cost volume filtering.} An intersection of two arrows indicates concatenation of two feature maps. The third number indicates the number of output channels. The second numbers are kernel sizes.}
    \label{fig:3d_unet}
\end{figure}
\subsection{U-Net for Cost Volume Filtering}
We use a U-Net~\cite{Ronneberger15UNet} with 2D convolutions for cost volume filtering, whose network architecture is shown in Fig.~\ref{fig:3d_unet}. We insert an ASPP (Atrous Spatial Pyramid Pooling) module~\cite{chen2018deeplab} in the bottleneck, where the dilation rates are 2, 4, and 8. It takes the dilated cost volumes with a shape of $C'\times\frac{H}{8}\times\frac{W}{8}$,\footnote{We omit the batch index for simplicity here.} where $C'=D\times C\times U\times V$. $D$ is the number of dilation rates ($D=7$ in this paper), $C$ is the output dimension of the similarity of two feature vectors ($C=1$ in this paper), and $U=V=2k+1$ are the neighborhood dimensions with $k$ being the neighborhood radius for the cost volume ($k=4$ in this paper so $U=V=9$). $H$ and $W$ are the height and width of input images, respectively. The output of the U-Net has a shape of $D'\times\frac{H}{8}\times\frac{W}{8}$, where $D'=D\times U\times V$.

\section{Visual Results of Optical Flow}
\begin{figure*}
\centering
\renewcommand{\tabcolsep}{0.8pt}
\newcommand{\loadFig}[1]{\includegraphics[width=0.19\linewidth]{#1}}
\begin{tabular}{ccccc}
\loadFig{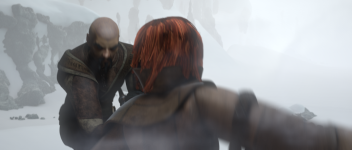} &
\loadFig{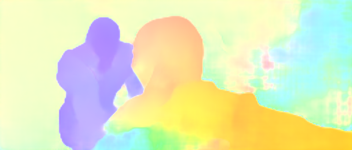} &
\loadFig{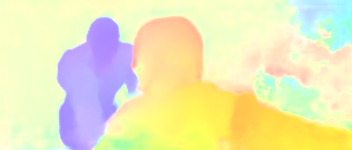} &
\loadFig{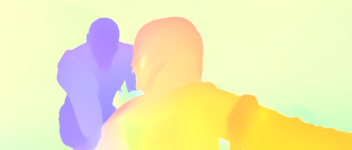} &
\loadFig{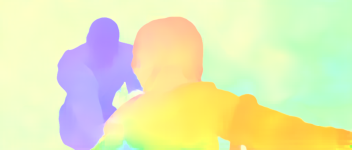} \\
\loadFig{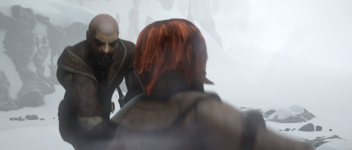} &
\loadFig{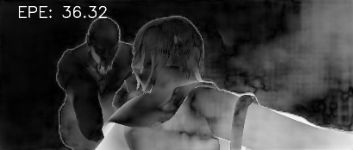} &
\loadFig{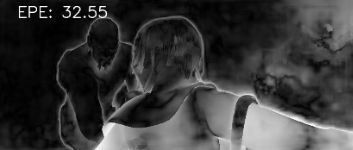} &
\loadFig{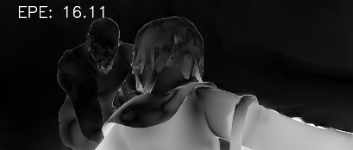} &
\loadFig{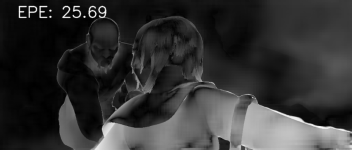} \\

\loadFig{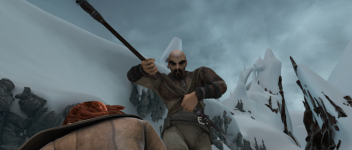} &
\loadFig{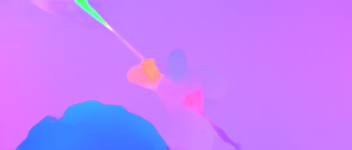} &
\loadFig{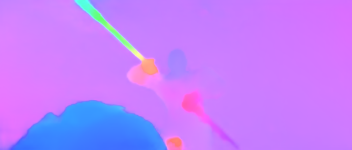} &
\loadFig{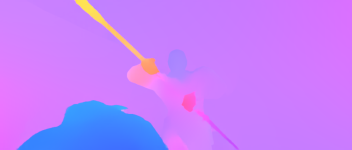} &
\loadFig{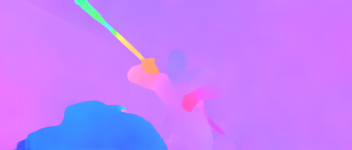} \\
\loadFig{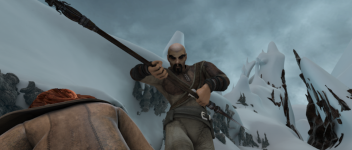} &
\loadFig{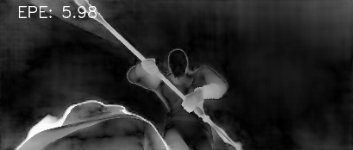} &
\loadFig{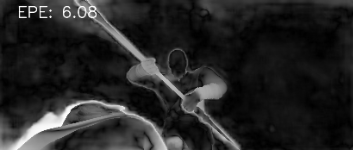} &
\loadFig{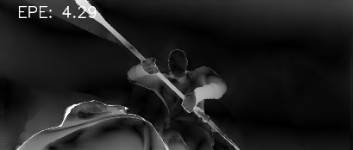} &
\loadFig{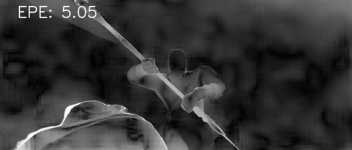} \\

\loadFig{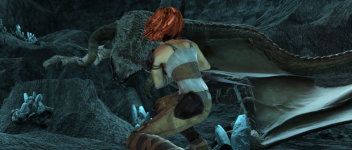} &
\loadFig{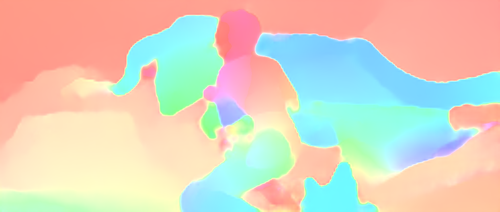} &
\loadFig{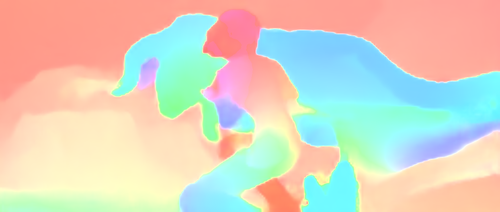} &
\loadFig{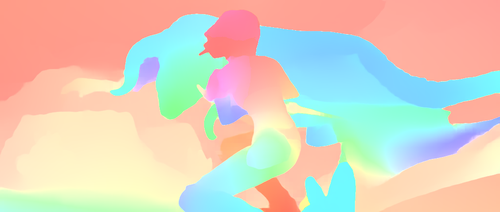} &
\loadFig{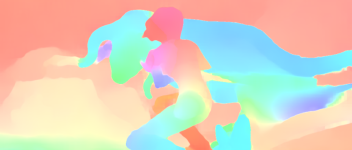} \\
\loadFig{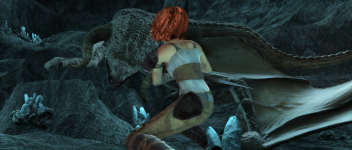} &
\loadFig{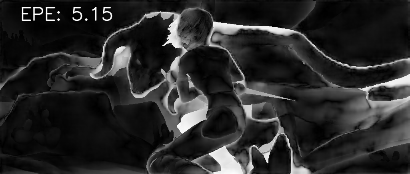} &
\loadFig{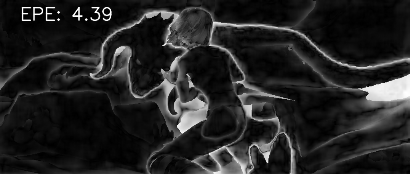} &
\loadFig{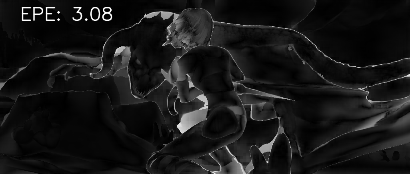} &
\loadFig{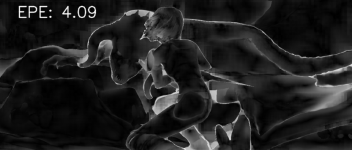} \\

\loadFig{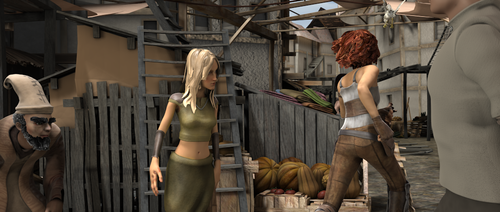} &
\loadFig{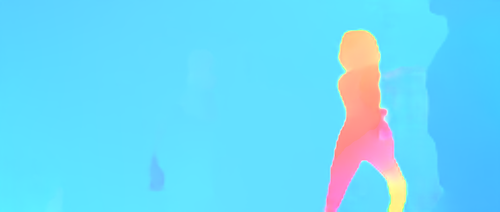} &
\loadFig{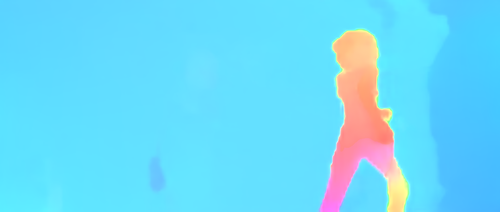} &
\loadFig{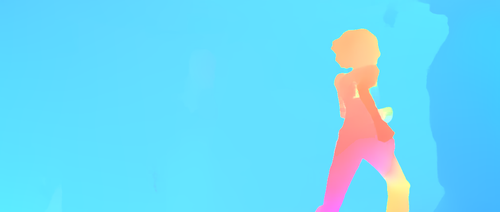} &
\loadFig{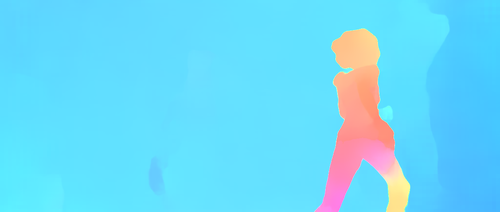} \\
\loadFig{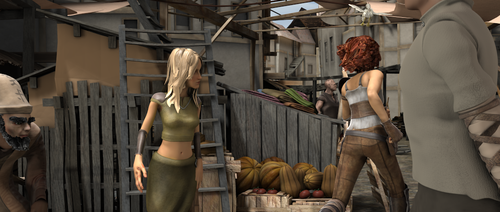} &
\loadFig{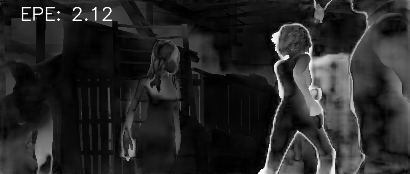} &
\loadFig{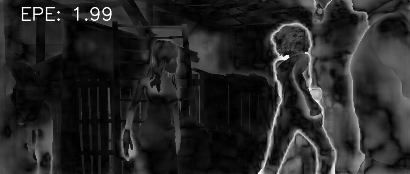} &
\loadFig{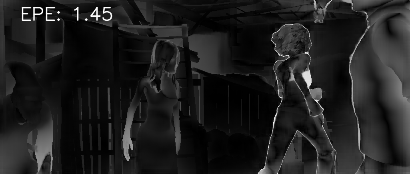} &
\loadFig{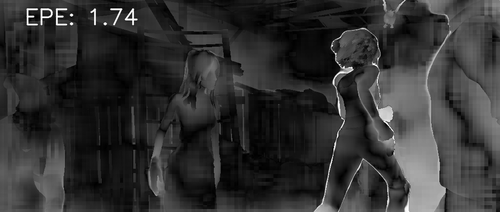} \\

\loadFig{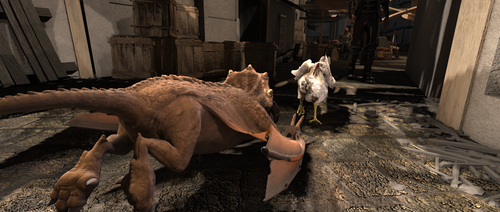} &
\loadFig{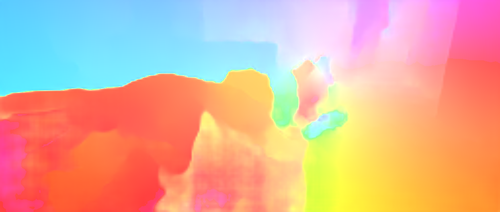} &
\loadFig{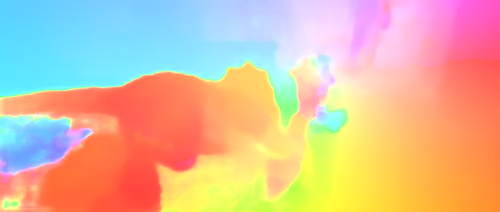} &
\loadFig{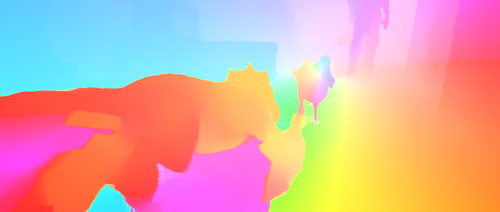} &
\loadFig{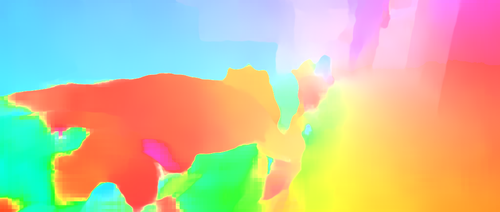} \\
\loadFig{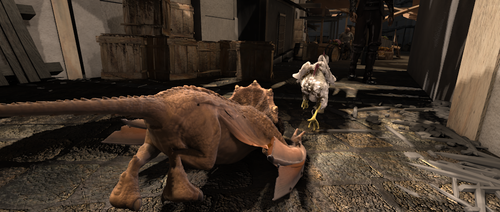} &
\loadFig{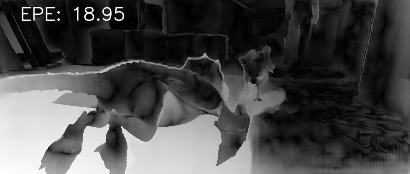} &
\loadFig{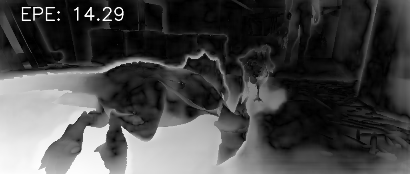} &
\loadFig{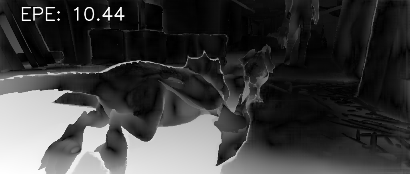} &
\loadFig{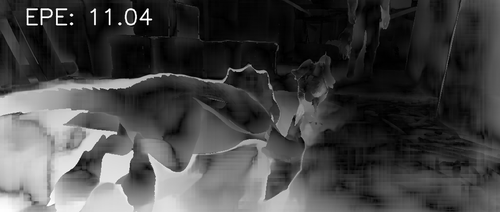} \\

\loadFig{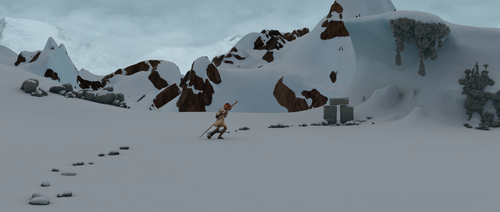} &
\loadFig{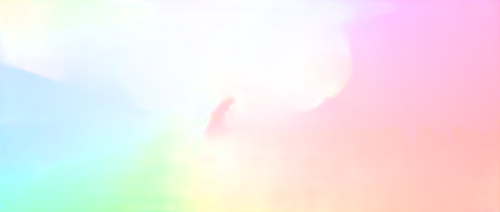} &
\loadFig{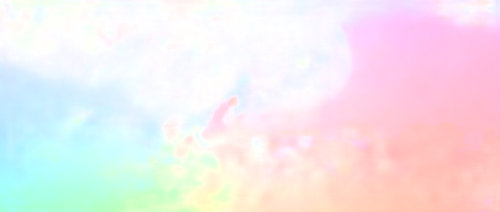} &
\loadFig{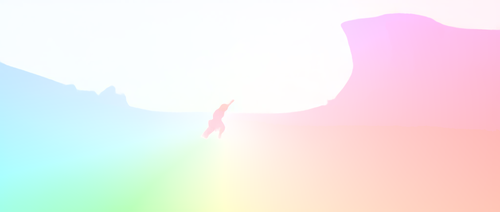} &
\loadFig{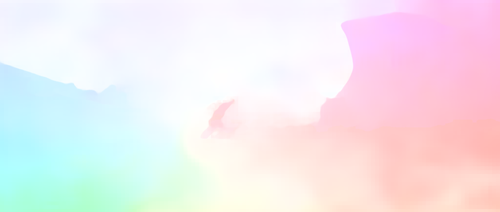} \\
\loadFig{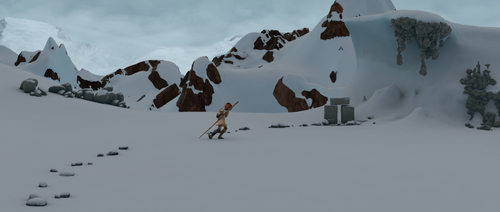} &
\loadFig{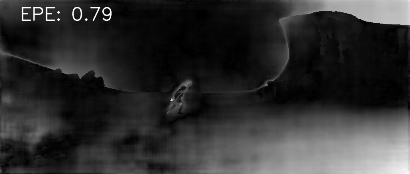} &
\loadFig{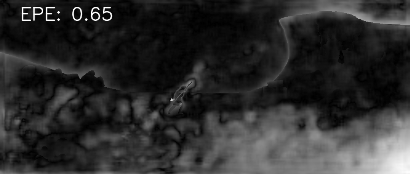} &
\loadFig{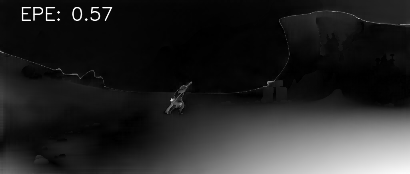} &
\loadFig{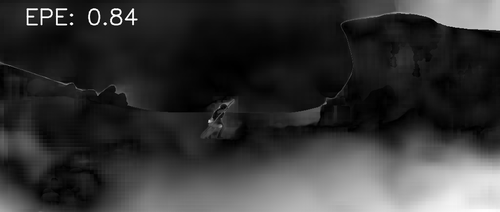} \\

\loadFig{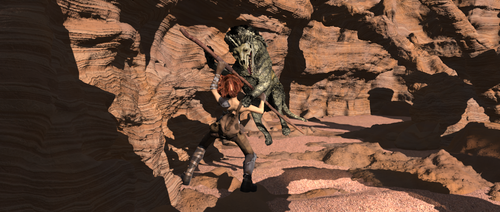} &
\loadFig{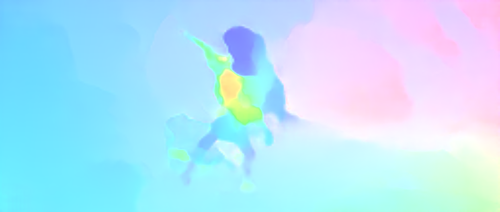} &
\loadFig{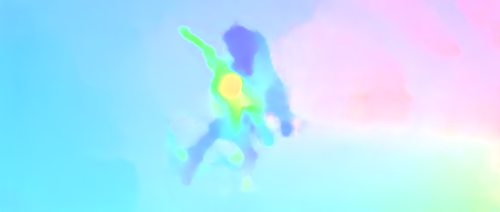} &
\loadFig{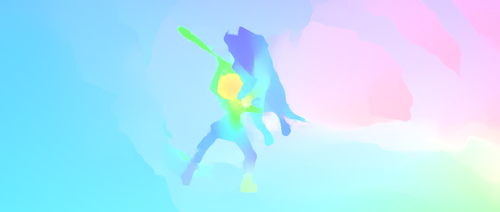} &
\loadFig{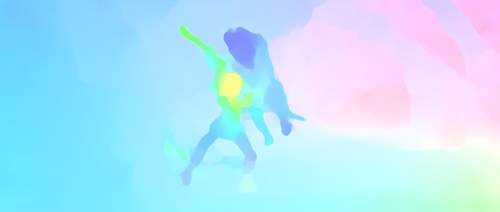} \\
\loadFig{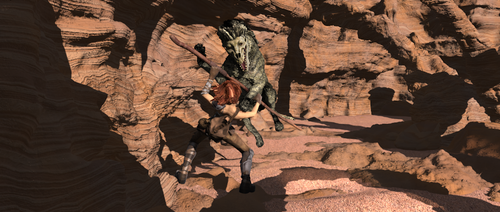} &
\loadFig{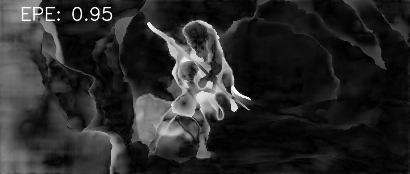} &
\loadFig{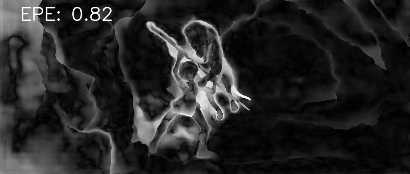} &
\loadFig{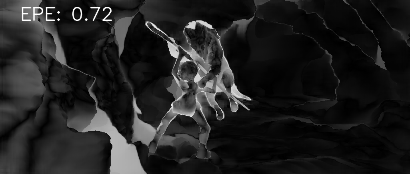} &
\loadFig{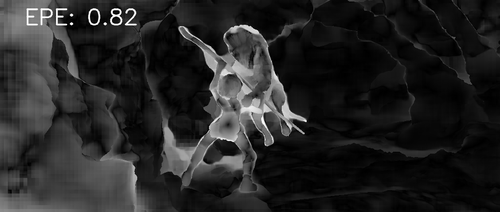} \\

(a) input images & (b) PWCNet~\cite{sun2018pwc} & (c) VCN~\cite{yang2019volumetric} & (d) RAFT~\cite{teed2020raft} & (e) our DCVNet \\
\end{tabular}
\caption{\textbf{Visual comparison of optical flow estimations on Sintel}. For each method, we show colorized optical flow and error maps (obtained from online servers). For the error maps, white indicates large error while black means small error. Best viewed in color.}
\label{fig:res_sintel_part_1}
\end{figure*}

\begin{figure*}
\centering
\renewcommand{\tabcolsep}{0.8pt}
\newcommand{\loadFig}[1]{\includegraphics[width=0.19\linewidth]{#1}}
\begin{tabular}{ccccc}
\loadFig{figs/supp_mat_v2/sintel/ambush_1_0021.png} &
\loadFig{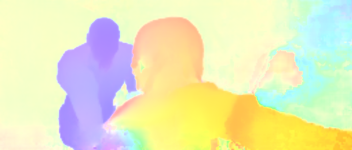} &
\loadFig{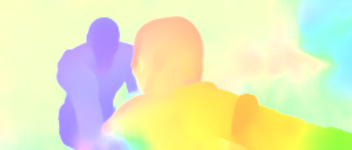} &
\loadFig{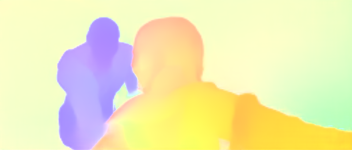} &
\loadFig{figs/supp_mat_v2/sintel/ambush_1_dcvnet.png} \\
\loadFig{figs/supp_mat_v2/sintel/ambush_1_0022.png} &
\loadFig{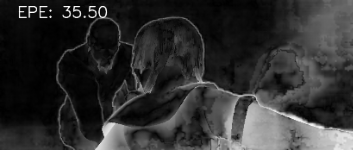} &
\loadFig{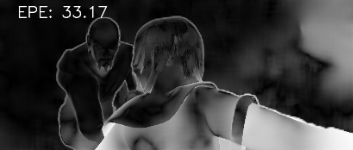} &
\loadFig{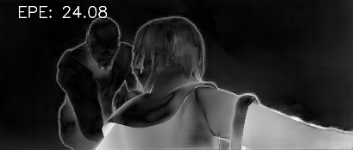} &
\loadFig{figs/supp_mat_v2/sintel/ambush_1_dcvnet_error.png} \\

\loadFig{figs/supp_mat_v2/sintel/ambush_3_0025.png} &
\loadFig{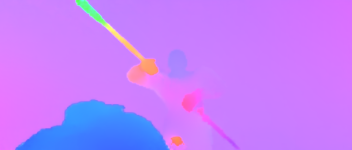} &
\loadFig{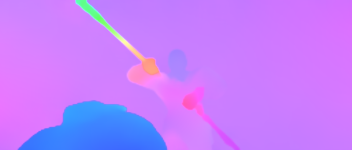} &
\loadFig{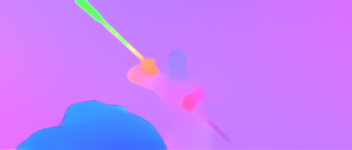} &
\loadFig{figs/supp_mat_v2/sintel/ambush_3_dcvnet.png} \\
\loadFig{figs/supp_mat_v2/sintel/ambush_3_0026.png} &
\loadFig{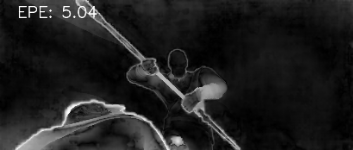} &
\loadFig{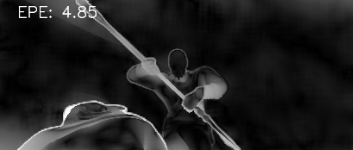} &
\loadFig{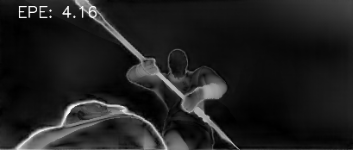} &
\loadFig{figs/supp_mat_v2/sintel/ambush_3_dcvnet_error.png} \\

\loadFig{figs/supp_mat_v2/sintel/cave_3_0016.png} &
\loadFig{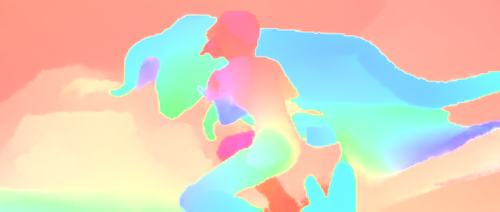} &
\loadFig{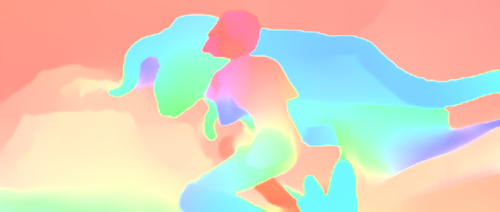} &
\loadFig{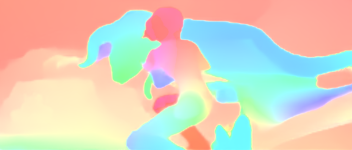} &
\loadFig{figs/supp_mat_v2/sintel/cave_3_dcvnet.png} \\
\loadFig{figs/supp_mat_v2/sintel/cave_3_0017.png} &
\loadFig{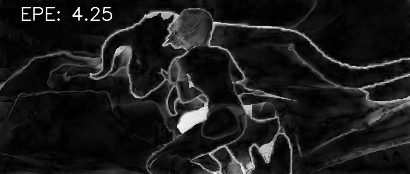} &
\loadFig{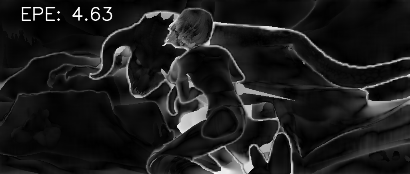} &
\loadFig{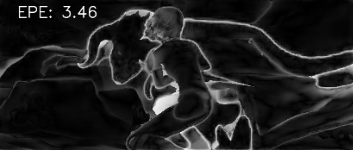} &
\loadFig{figs/supp_mat_v2/sintel/cave_3_dcvnet_error.png} \\

\loadFig{figs/supp_mat_v2/sintel/market_1_0018.png} &
\loadFig{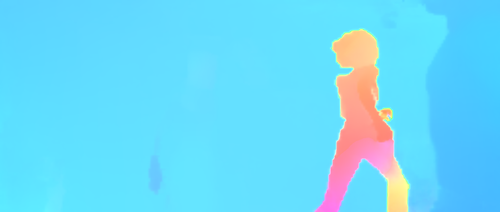} &
\loadFig{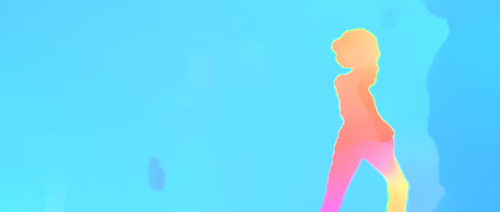} &
\loadFig{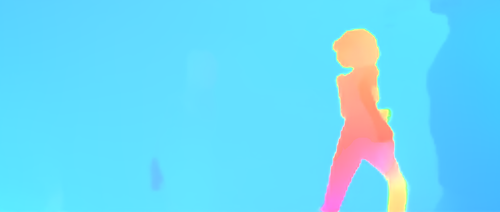} &
\loadFig{figs/supp_mat_v2/sintel/market_1_dcvnet.png} \\
\loadFig{figs/supp_mat_v2/sintel/market_1_0019.png} &
\loadFig{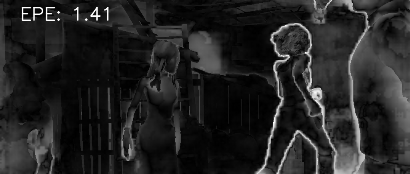} &
\loadFig{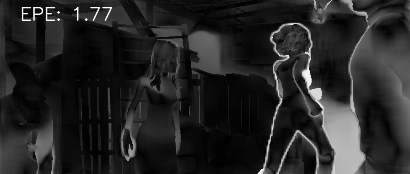} &
\loadFig{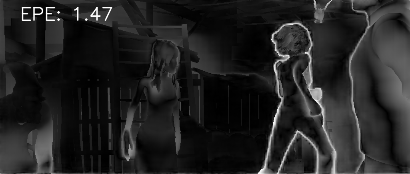} &
\loadFig{figs/supp_mat_v2/sintel/market_1_dcvnet_error.png} \\

\loadFig{figs/supp_mat_v2/sintel/market_4_0047.png} &
\loadFig{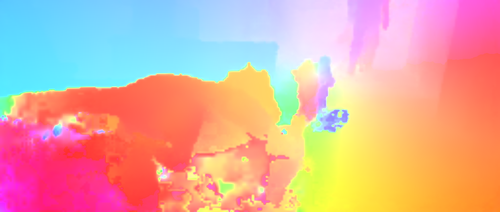} &
\loadFig{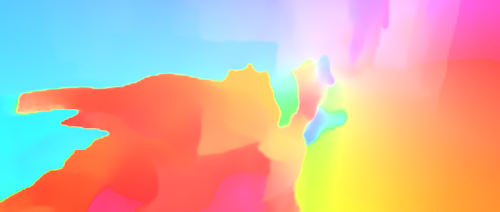} &
\loadFig{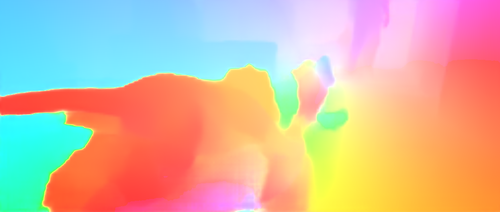} &
\loadFig{figs/supp_mat_v2/sintel/market_4_dcvnet.png} \\
\loadFig{figs/supp_mat_v2/sintel/market_4_0048.png} &
\loadFig{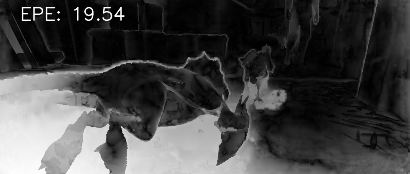} &
\loadFig{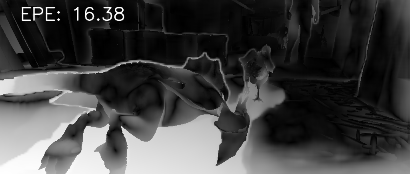} &
\loadFig{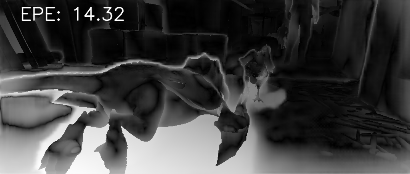} &
\loadFig{figs/supp_mat_v2/sintel/market_4_dcvnet_error.png} \\

\loadFig{figs/supp_mat_v2/sintel/mountain_2_0048.png} &
\loadFig{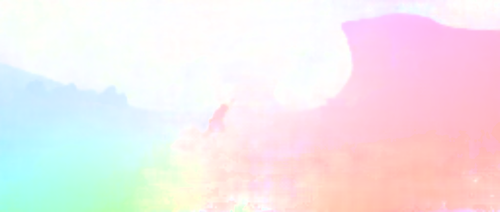} &
\loadFig{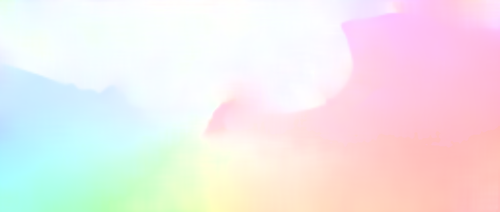} &
\loadFig{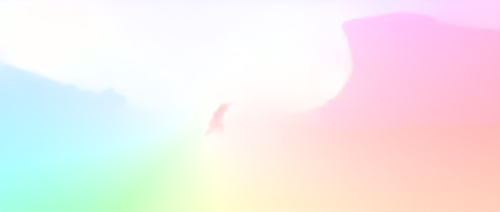} &
\loadFig{figs/supp_mat_v2/sintel/mountain_2_dcvnet.png} \\
\loadFig{figs/supp_mat_v2/sintel/mountain_2_0049.png} &
\loadFig{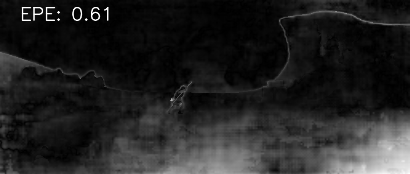} &
\loadFig{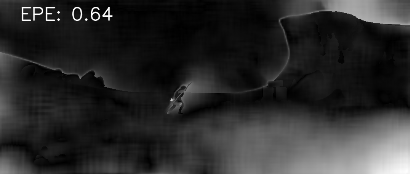} &
\loadFig{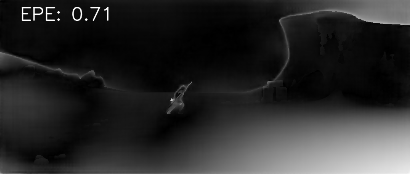} &
\loadFig{figs/supp_mat_v2/sintel/mountain_2_dcvnet_error.png} \\

\loadFig{figs/supp_mat_v2/sintel/tiger_0040.png} &
\loadFig{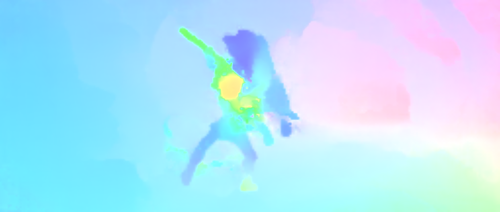} &
\loadFig{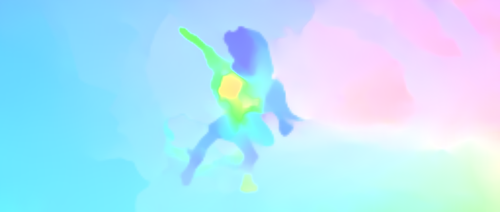} &
\loadFig{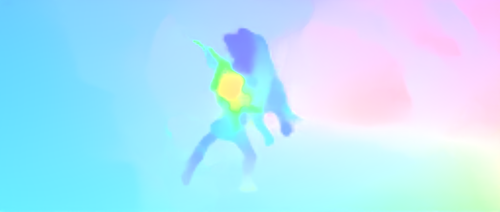} &
\loadFig{figs/supp_mat_v2/sintel/tiger_dcvnet.png} \\
\loadFig{figs/supp_mat_v2/sintel/tiger_0041.png} &
\loadFig{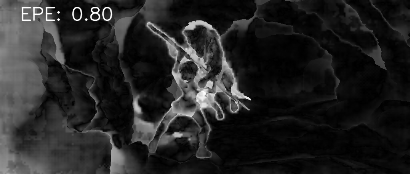} &
\loadFig{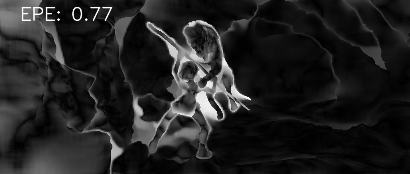} &
\loadFig{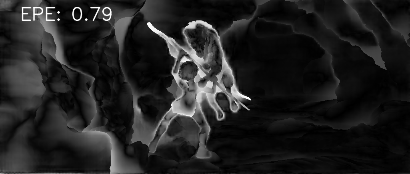} &
\loadFig{figs/supp_mat_v2/sintel/tiger_dcvnet_error.png} \\

(a) input images & (b) HD3~\cite{yin2019hierarchical} & (c) LiteFlowNet3~\cite{hui2020liteflownet3} & (d) DICL~\cite{wang2020displacement} & (e) our DCVNet \\
\end{tabular}
\caption{\textbf{Visual comparison of optical flow estimations on Sintel}. For each method, we show colorized optical flow and error maps (obtained from online servers). For the error maps, white indicates large error while black means small error. Best viewed in color.}
\label{fig:res_sintel_part_2}
\end{figure*}

\begin{figure*}
\centering
\renewcommand{\tabcolsep}{0.8pt}
\newcommand{\loadFig}[1]{\includegraphics[width=0.19\linewidth]{#1}}
\begin{tabular}{ccccc}
\loadFig{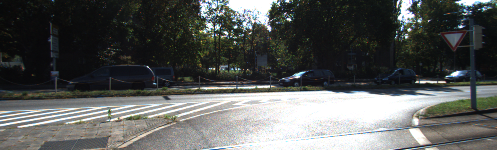} & 
\loadFig{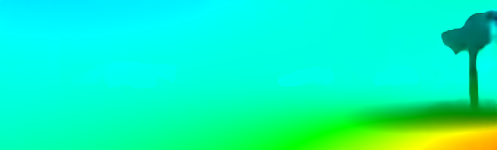} & 
\loadFig{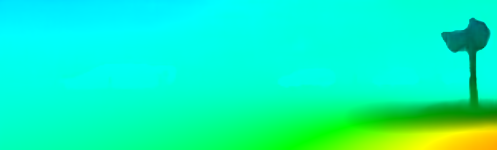} & 
\loadFig{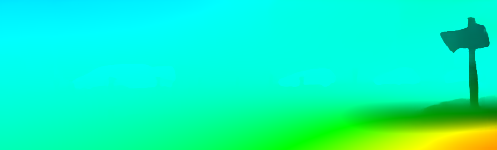} & 
\loadFig{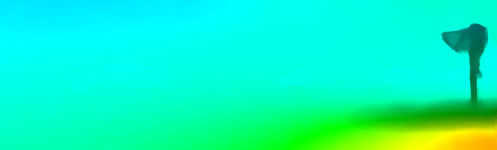} \\ 
\loadFig{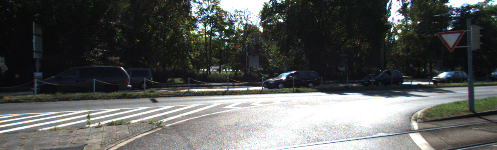} & 
\loadFig{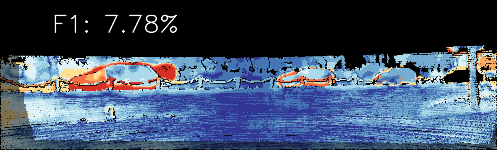} & 
\loadFig{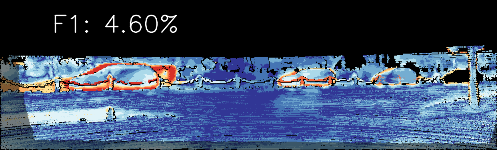} & 
\loadFig{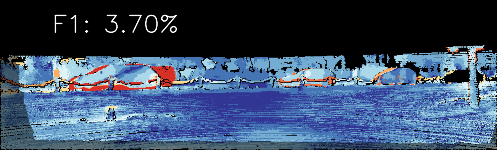} & 
\loadFig{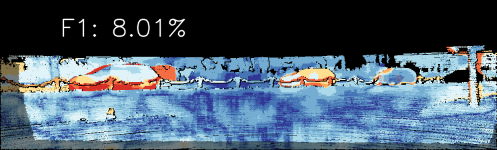} \\ 

\loadFig{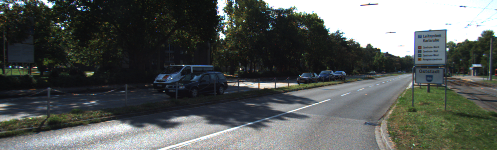} & 
\loadFig{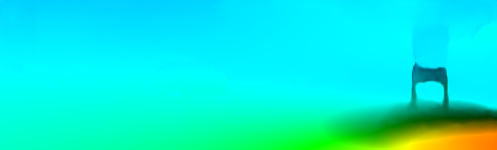} & 
\loadFig{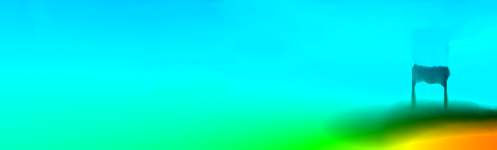} & 
\loadFig{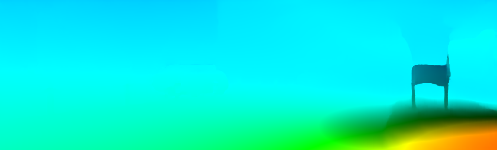} & 
\loadFig{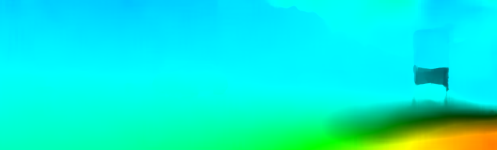} \\ 
\loadFig{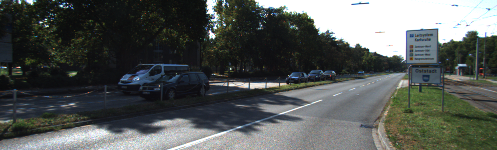} & 
\loadFig{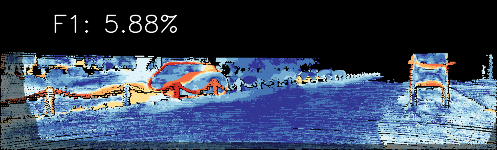} & 
\loadFig{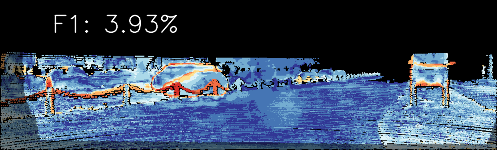} & 
\loadFig{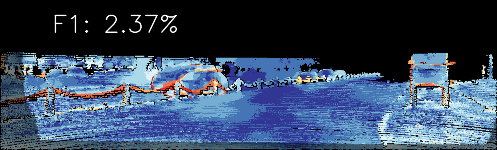} & 
\loadFig{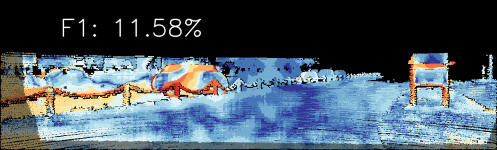} \\ 

\loadFig{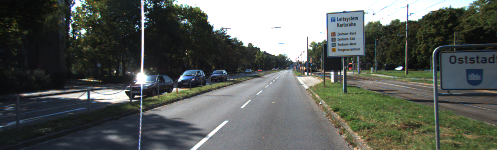} & 
\loadFig{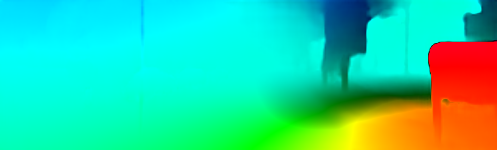} & 
\loadFig{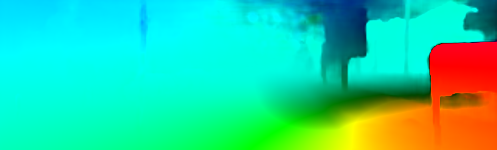} & 
\loadFig{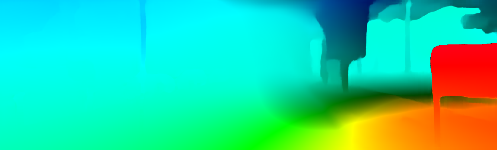} & 
\loadFig{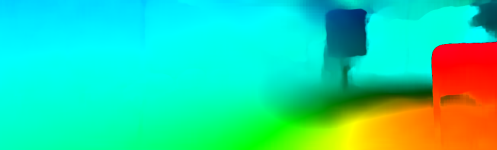} \\ 
\loadFig{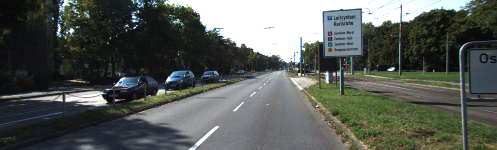} & 
\loadFig{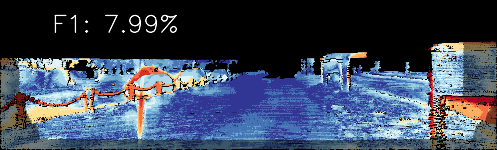} & 
\loadFig{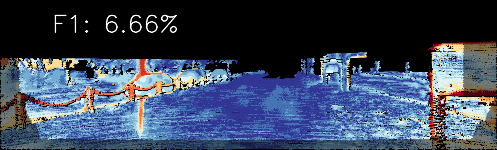} & 
\loadFig{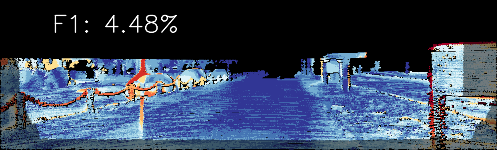} & 
\loadFig{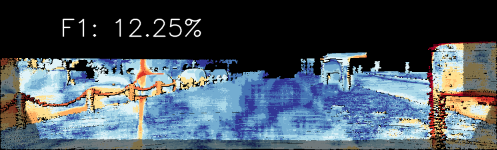} \\ 

\loadFig{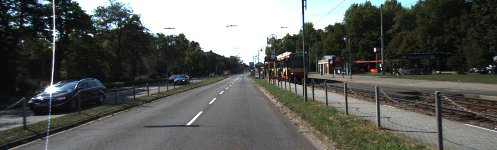} & 
\loadFig{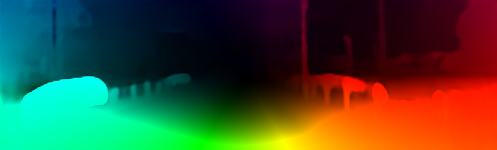} & 
\loadFig{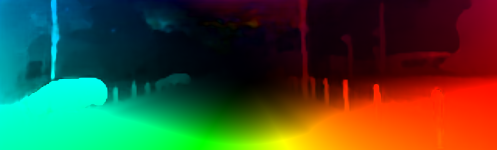} & 
\loadFig{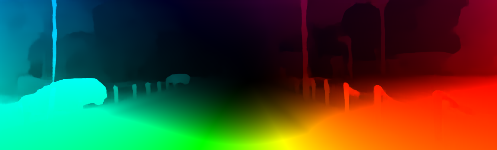} & 
\loadFig{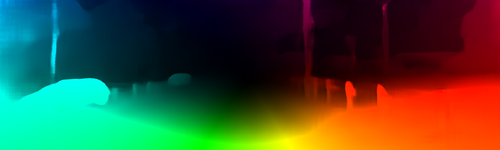} \\ 
\loadFig{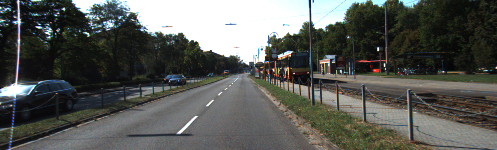} & 
\loadFig{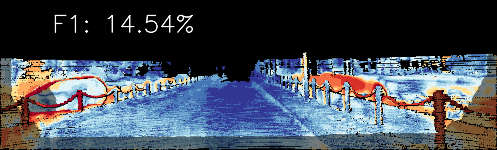} & 
\loadFig{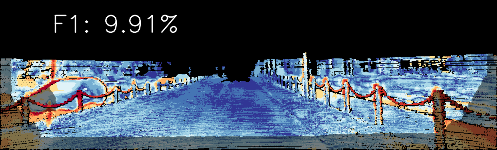} & 
\loadFig{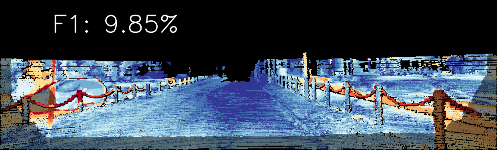} & 
\loadFig{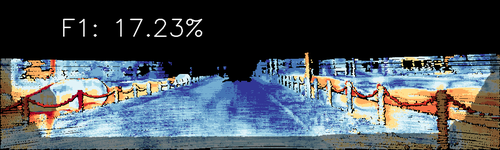} \\ 

\loadFig{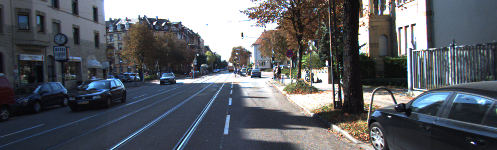} & 
\loadFig{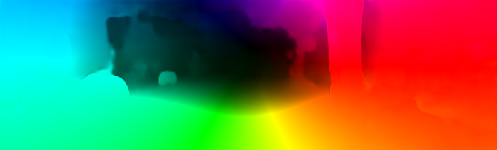} & 
\loadFig{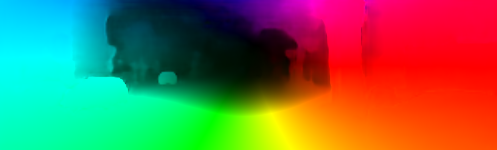} & 
\loadFig{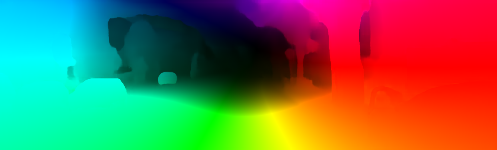} & 
\loadFig{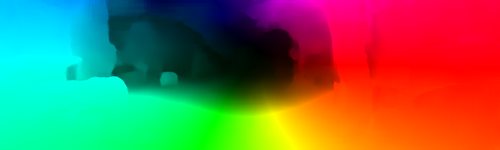} \\ 
\loadFig{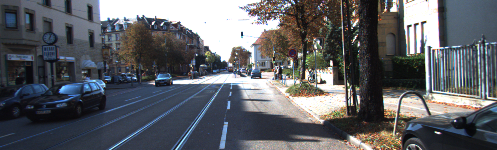} & 
\loadFig{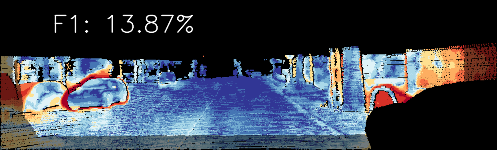} & 
\loadFig{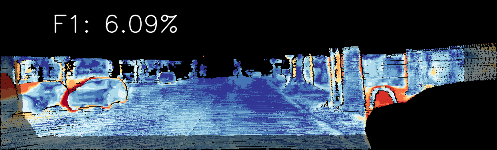} & 
\loadFig{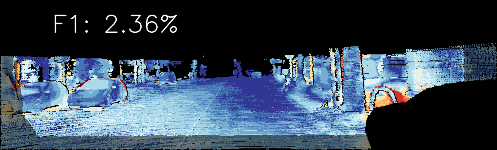} & 
\loadFig{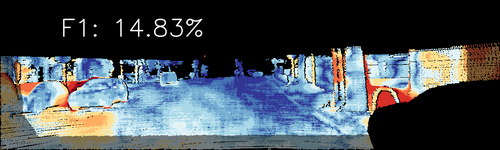} \\ 

\loadFig{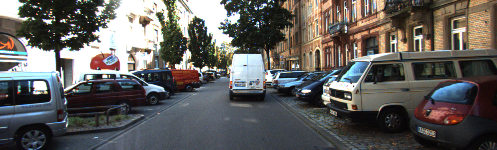} & 
\loadFig{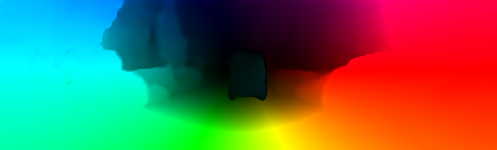} & 
\loadFig{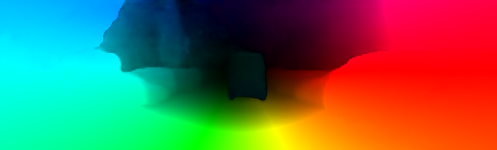} & 
\loadFig{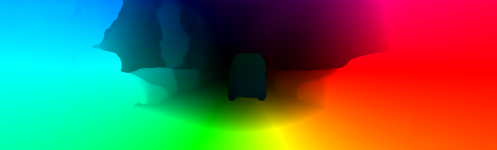} & 
\loadFig{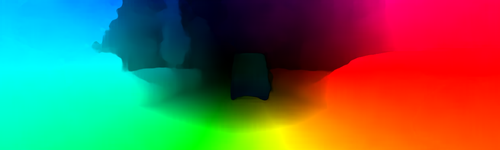} \\ 
\loadFig{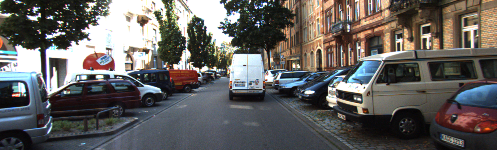} & 
\loadFig{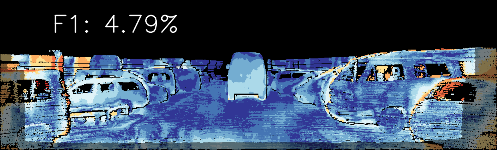} & 
\loadFig{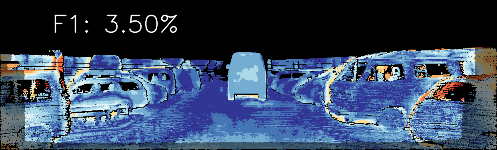} & 
\loadFig{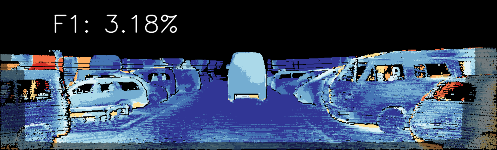} & 
\loadFig{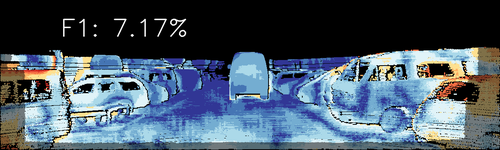} \\ 

\loadFig{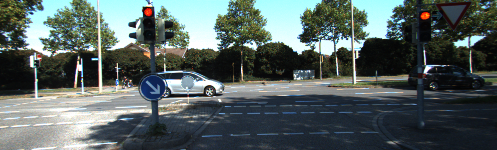} & 
\loadFig{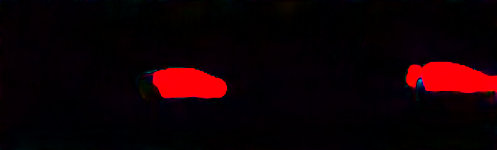} & 
\loadFig{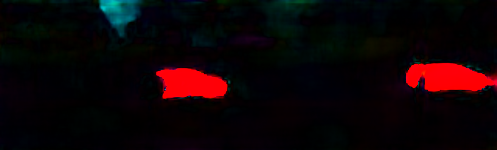} & 
\loadFig{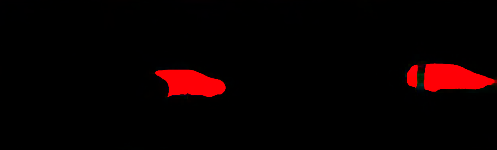} & 
\loadFig{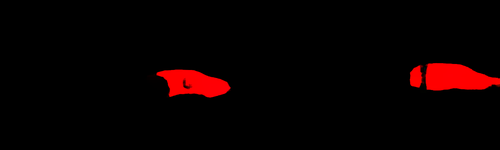} \\ 
\loadFig{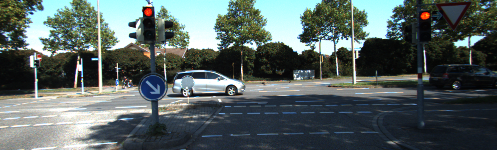} & 
\loadFig{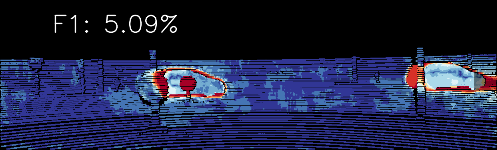} & 
\loadFig{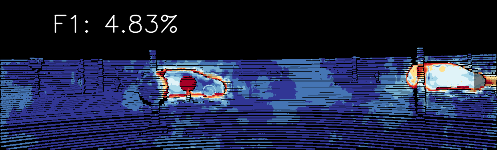} & 
\loadFig{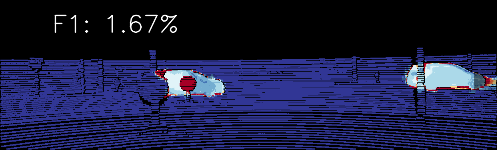} & 
\loadFig{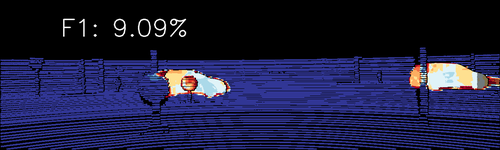} \\ 

\loadFig{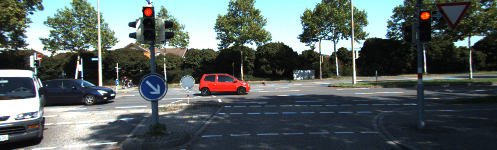} & 
\loadFig{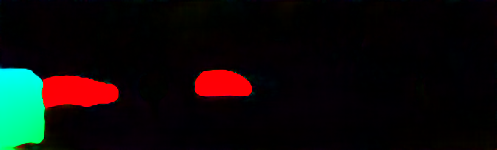} & 
\loadFig{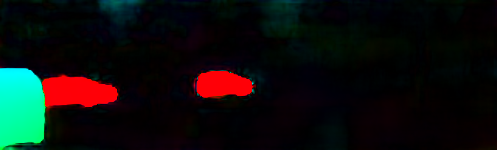} & 
\loadFig{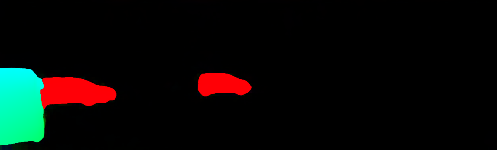} & 
\loadFig{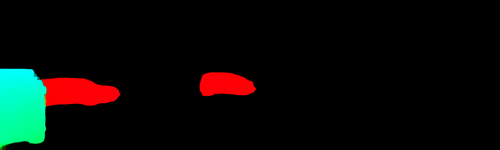} \\ 
\loadFig{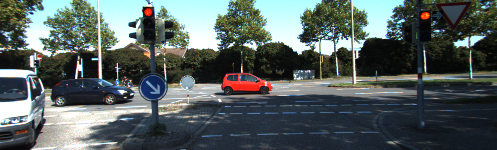} & 
\loadFig{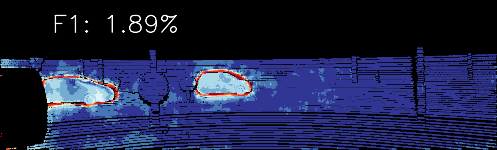} & 
\loadFig{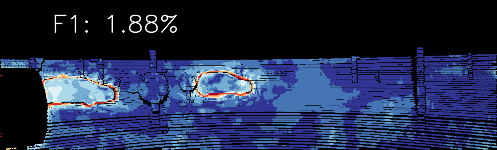} & 
\loadFig{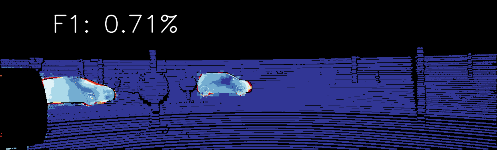} & 
\loadFig{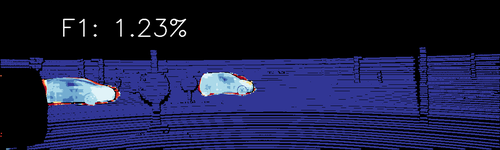} \\ 

\loadFig{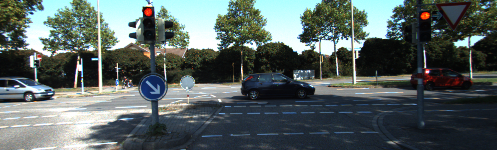} & 
\loadFig{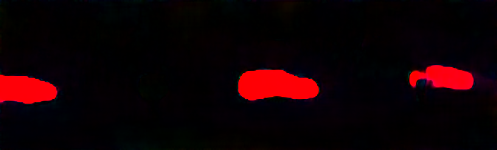} & 
\loadFig{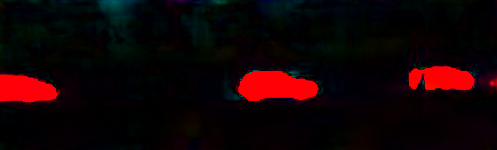} & 
\loadFig{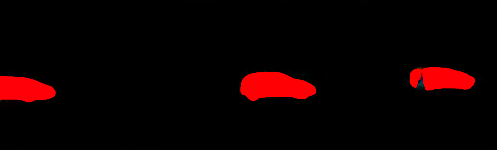} & 
\loadFig{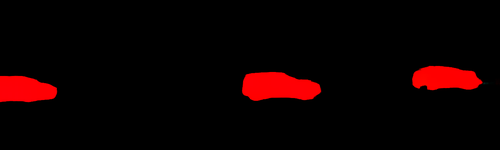} \\ 
\loadFig{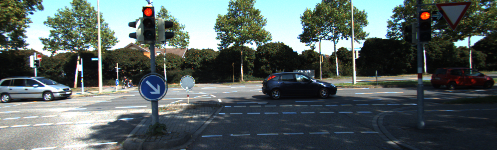} & 
\loadFig{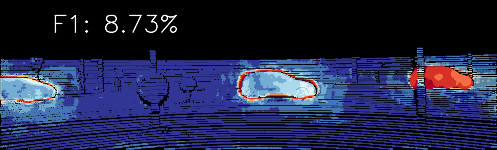} & 
\loadFig{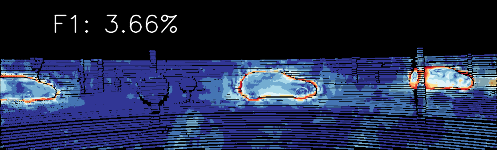} & 
\loadFig{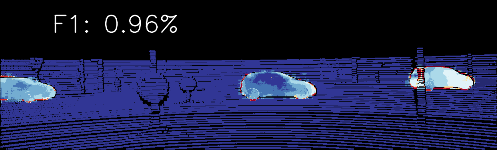} & 
\loadFig{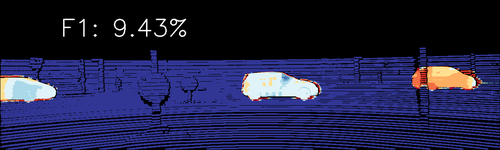} \\  

(a) input images & (b) PWCNet~\cite{sun2018pwc} & (c) VCN~\cite{yang2019volumetric} & (d) RAFT~\cite{teed2020raft} & (e) our DCVNet \\
\end{tabular}
\caption{\textbf{Visual comparison of optical flow estimations on KITTI 2015}. For each method, we show colorized optical flow and error maps (obtained from online servers). For the error maps, red indicates large error while blue means small error. Best viewed in color.}
\label{fig:res_k15_part_1}
\end{figure*}

\begin{figure*}
\centering
\renewcommand{\tabcolsep}{0.8pt}
\newcommand{\loadFig}[1]{\includegraphics[width=0.19\linewidth]{#1}}
\begin{tabular}{ccccc}
\loadFig{figs/supp_mat_v2/k15/000000_10.png} & 
\loadFig{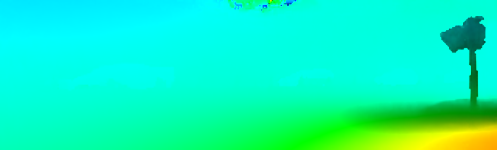} & 
\loadFig{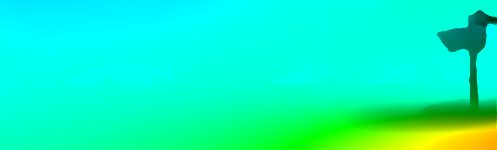} & 
\loadFig{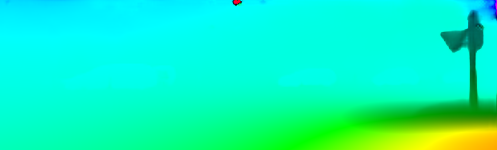} & 
\loadFig{figs/supp_mat_v2/k15/000000_10_dcvnet.png} \\ 
\loadFig{figs/supp_mat_v2/k15/000000_11.png} & 
\loadFig{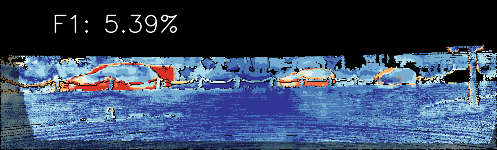} & 
\loadFig{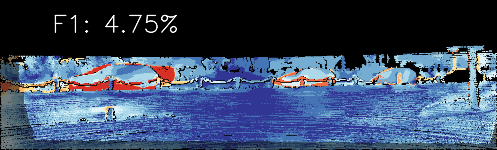} & 
\loadFig{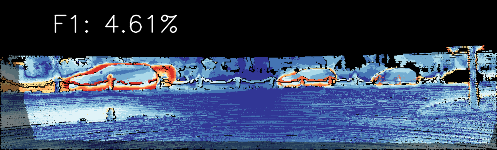} & 
\loadFig{figs/supp_mat_v2/k15/000000_10_dcvnet_error.png} \\ 

\loadFig{figs/supp_mat_v2/k15/000001_10.png} & 
\loadFig{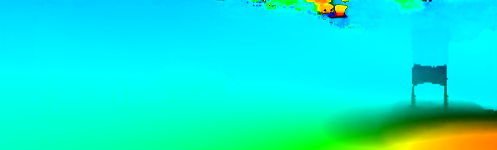} & 
\loadFig{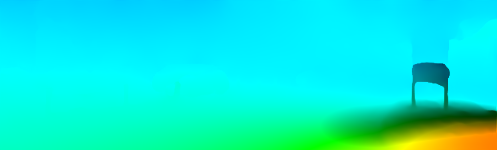} & 
\loadFig{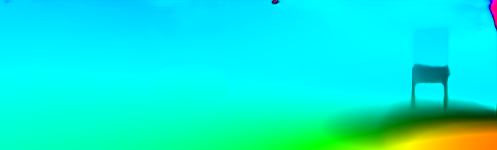} & 
\loadFig{figs/supp_mat_v2/k15/000001_10_dcvnet.png} \\ 
\loadFig{figs/supp_mat_v2/k15/000001_11.png} & 
\loadFig{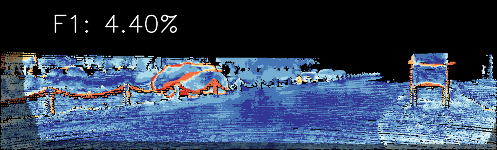} & 
\loadFig{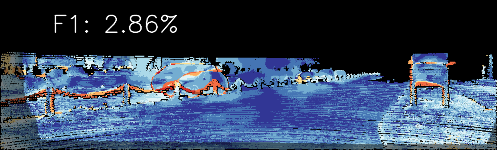} & 
\loadFig{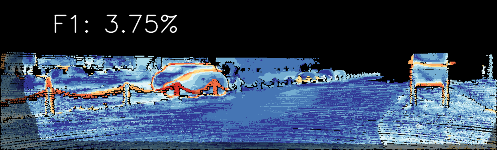} & 
\loadFig{figs/supp_mat_v2/k15/000001_10_dcvnet_error.png} \\ 

\loadFig{figs/supp_mat_v2/k15/000002_10.png} & 
\loadFig{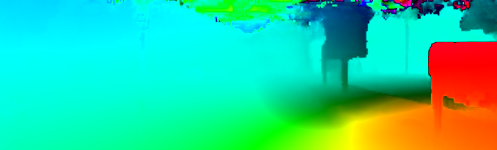} & 
\loadFig{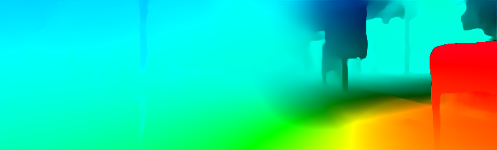} & 
\loadFig{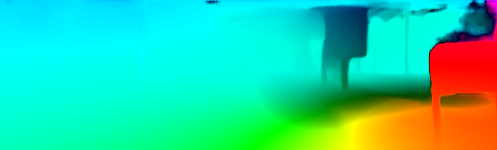} & 
\loadFig{figs/supp_mat_v2/k15/000002_10_dcvnet.png} \\ 
\loadFig{figs/supp_mat_v2/k15/000002_11.png} & 
\loadFig{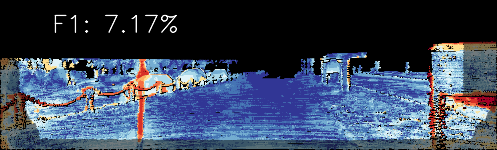} & 
\loadFig{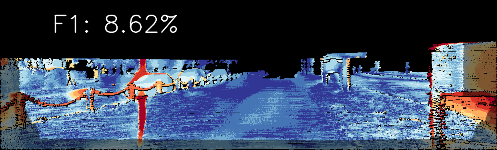} & 
\loadFig{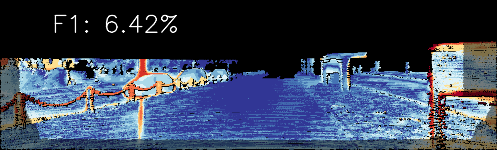} & 
\loadFig{figs/supp_mat_v2/k15/000002_10_dcvnet_error.png} \\ 

\loadFig{figs/supp_mat_v2/k15/000003_10.png} & 
\loadFig{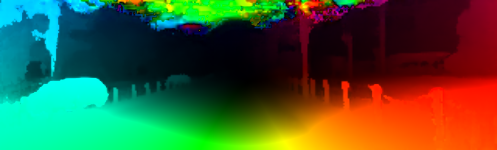} & 
\loadFig{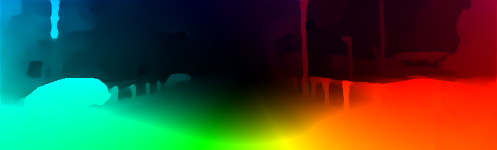} & 
\loadFig{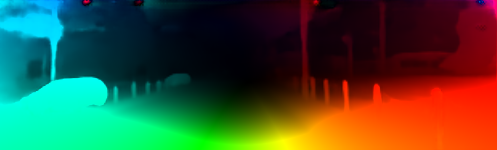} & 
\loadFig{figs/supp_mat_v2/k15/000003_10_dcvnet.png} \\ 
\loadFig{figs/supp_mat_v2/k15/000003_11.png} & 
\loadFig{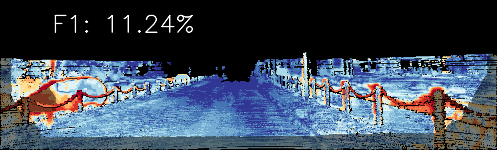} & 
\loadFig{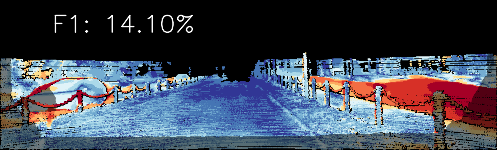} & 
\loadFig{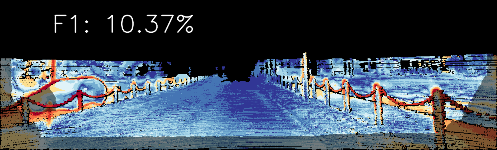} & 
\loadFig{figs/supp_mat_v2/k15/000003_10_dcvnet_error.png} \\ 

\loadFig{figs/supp_mat_v2/k15/000004_10.png} & 
\loadFig{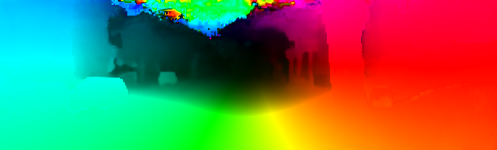} & 
\loadFig{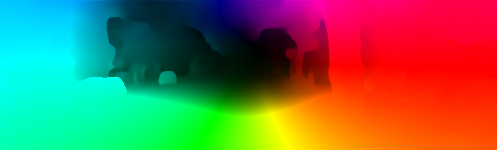} & 
\loadFig{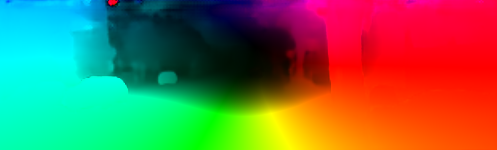} & 
\loadFig{figs/supp_mat_v2/k15/000004_10_dcvnet.png} \\ 
\loadFig{figs/supp_mat_v2/k15/000004_11.png} & 
\loadFig{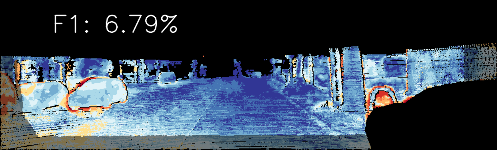} & 
\loadFig{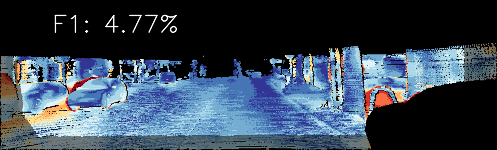} & 
\loadFig{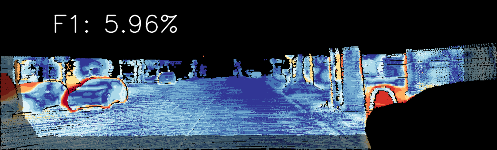} & 
\loadFig{figs/supp_mat_v2/k15/000004_10_dcvnet_error.png} \\ 

\loadFig{figs/supp_mat_v2/k15/000005_10.png} & 
\loadFig{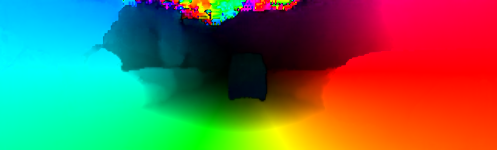} & 
\loadFig{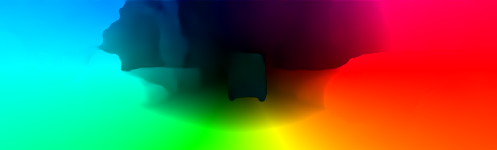} & 
\loadFig{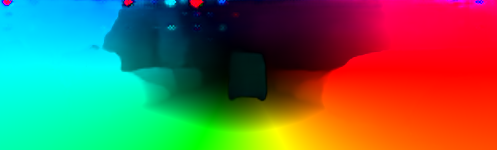} & 
\loadFig{figs/supp_mat_v2/k15/000005_10_dcvnet.png} \\ 
\loadFig{figs/supp_mat_v2/k15/000005_11.png} & 
\loadFig{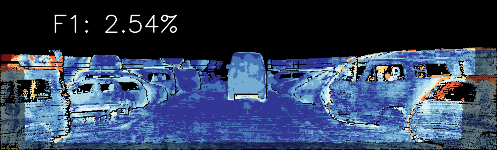} & 
\loadFig{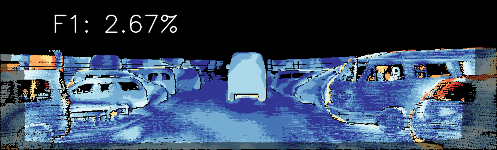} & 
\loadFig{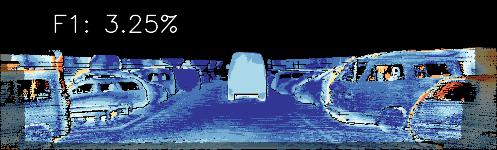} & 
\loadFig{figs/supp_mat_v2/k15/000005_10_dcvnet_error.png} \\ 

\loadFig{figs/supp_mat_v2/k15/000007_10.png} & 
\loadFig{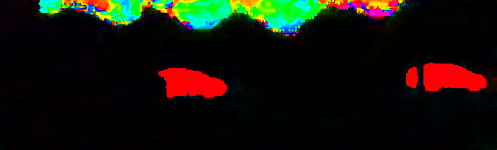} & 
\loadFig{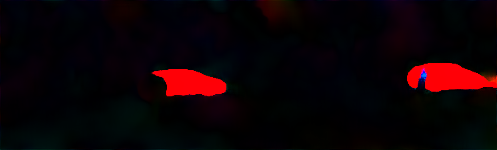} & 
\loadFig{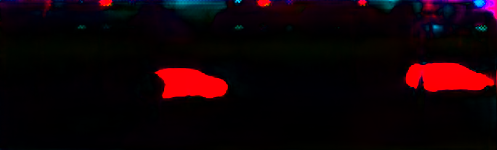} & 
\loadFig{figs/supp_mat_v2/k15/000007_10_dcvnet.png} \\ 
\loadFig{figs/supp_mat_v2/k15/000007_11.png} & 
\loadFig{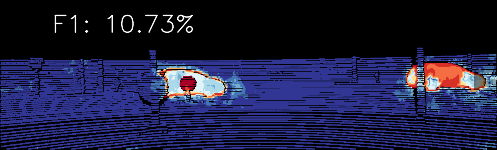} & 
\loadFig{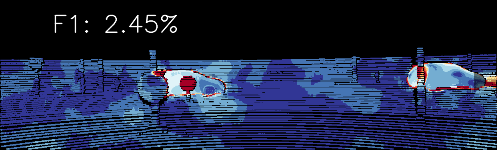} & 
\loadFig{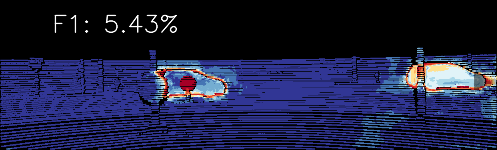} & 
\loadFig{figs/supp_mat_v2/k15/000007_10_dcvnet_error.png} \\ 

\loadFig{figs/supp_mat_v2/k15/000008_10.png} & 
\loadFig{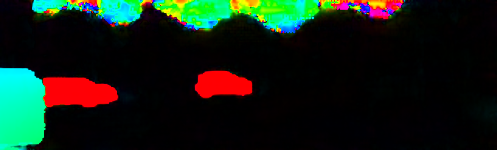} & 
\loadFig{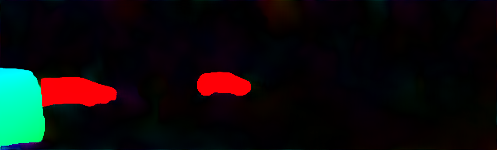} & 
\loadFig{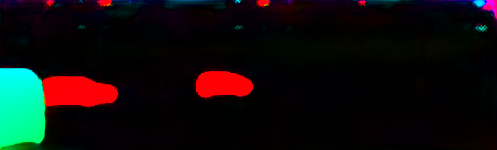} & 
\loadFig{figs/supp_mat_v2/k15/000008_10_dcvnet.png} \\ 
\loadFig{figs/supp_mat_v2/k15/000008_11.png} & 
\loadFig{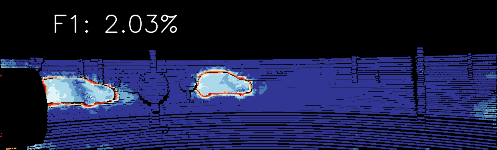} & 
\loadFig{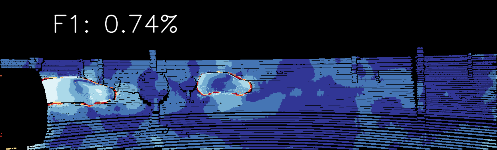} & 
\loadFig{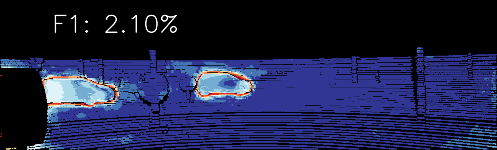} & 
\loadFig{figs/supp_mat_v2/k15/000008_10_dcvnet_error.png} \\ 

\loadFig{figs/supp_mat_v2/k15/000009_10.png} & 
\loadFig{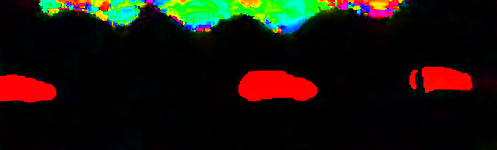} & 
\loadFig{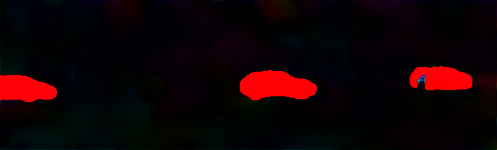} & 
\loadFig{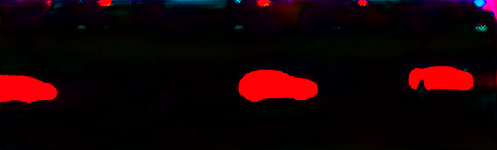} & 
\loadFig{figs/supp_mat_v2/k15/000009_10_dcvnet.png} \\ 
\loadFig{figs/supp_mat_v2/k15/000009_11.png} & 
\loadFig{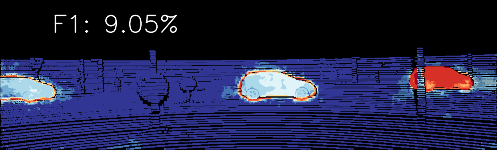} & 
\loadFig{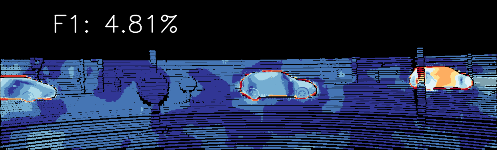} & 
\loadFig{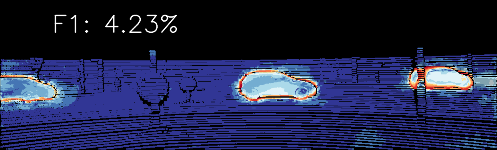} & 
\loadFig{figs/supp_mat_v2/k15/000009_10_dcvnet_error.png} \\ 

(a) input images & (b) HD3~\cite{yin2019hierarchical} & (c) LiteFlowNet3~\cite{hui2020liteflownet3} & (d) DICL~\cite{wang2020displacement} & (e) our DCVNet \\
\end{tabular}
\caption{\textbf{Visual comparison of optical flow estimations on KITTI 2015}. For each method, we show colorized optical flow and error maps (obtained from online servers). For the error maps, red indicates large error while blue means small error. Best viewed in color.}
\label{fig:res_k15_part_2}
\end{figure*}

\begin{figure*}
\centering
\renewcommand{\tabcolsep}{0.8pt}
\newcommand{\loadFig}[1]{\includegraphics[width=0.19\linewidth]{#1}}
\begin{tabular}{ccccc}
\loadFig{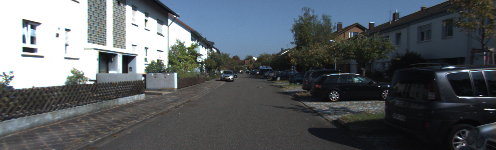} & 
\loadFig{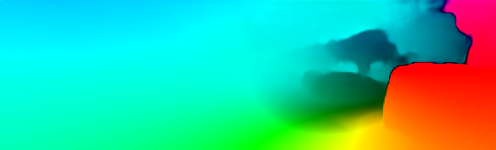} & 
\loadFig{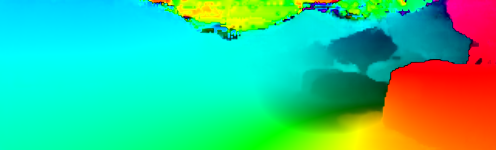} & 
\loadFig{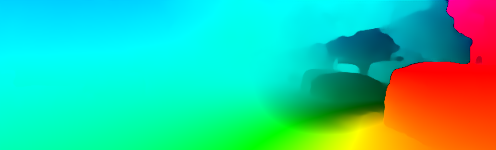} & 
\loadFig{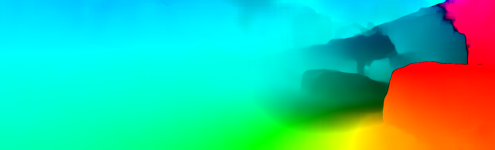} \\ 
\loadFig{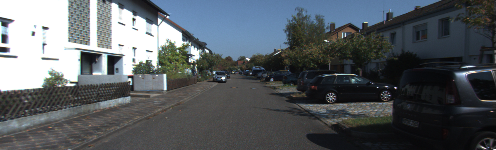} & 
\loadFig{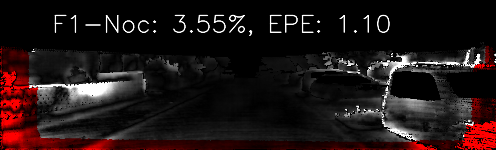} & 
\loadFig{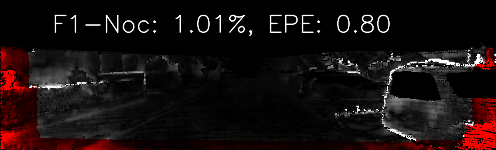} & 
\loadFig{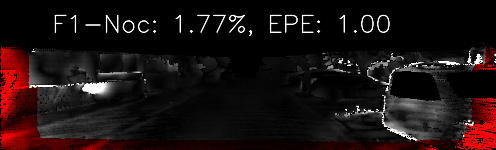} & 
\loadFig{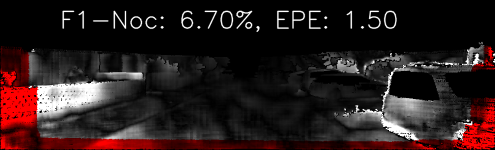} \\ 

\loadFig{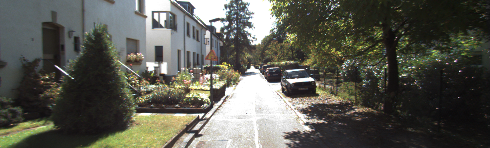} & 
\loadFig{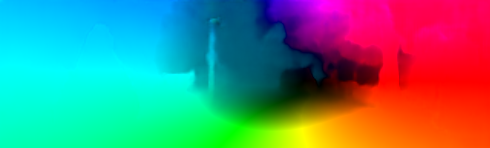} & 
\loadFig{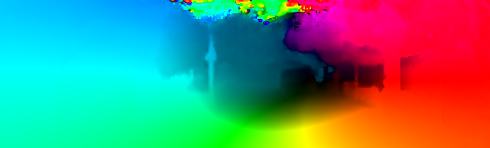} & 
\loadFig{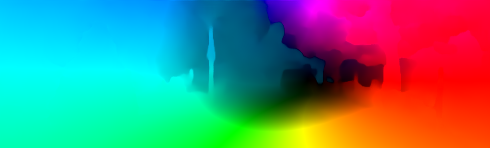} & 
\loadFig{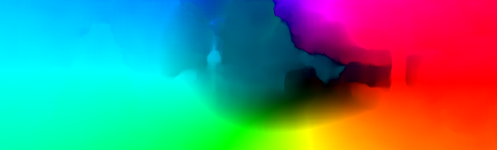} \\ 
\loadFig{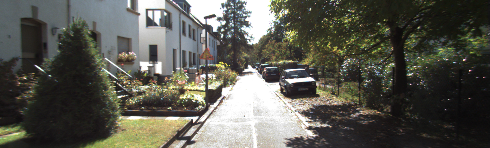} & 
\loadFig{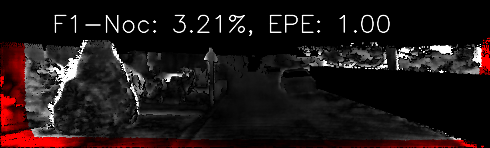} & 
\loadFig{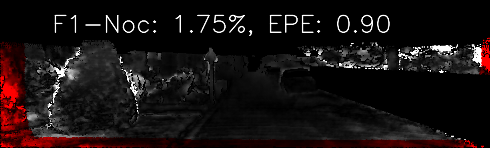} & 
\loadFig{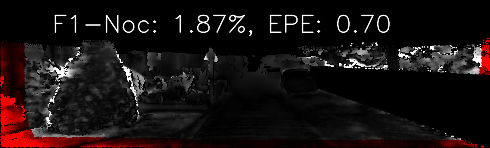} & 
\loadFig{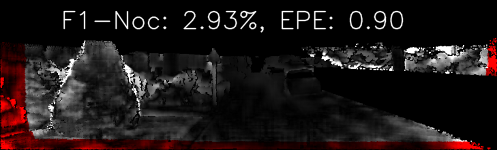} \\ 

\loadFig{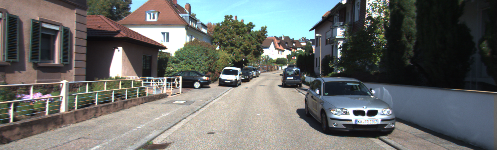} & 
\loadFig{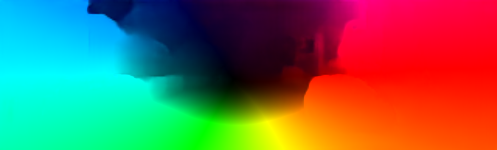} & 
\loadFig{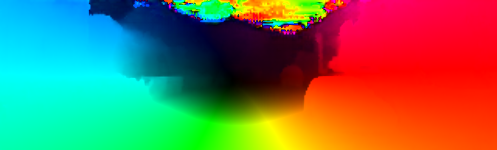} & 
\loadFig{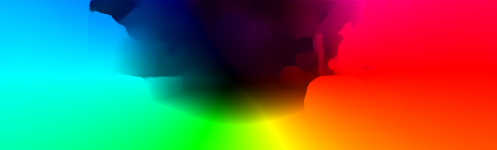} & 
\loadFig{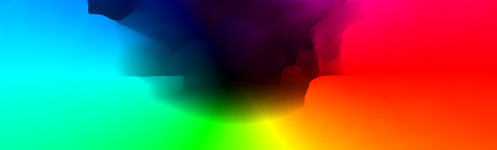} \\ 
\loadFig{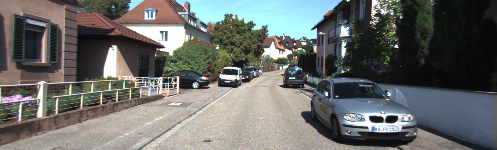} & 
\loadFig{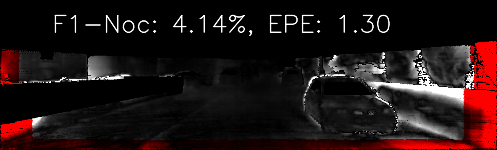} & 
\loadFig{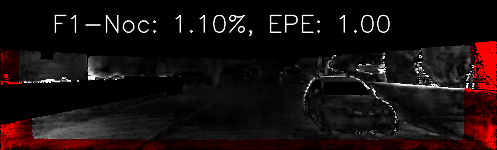} & 
\loadFig{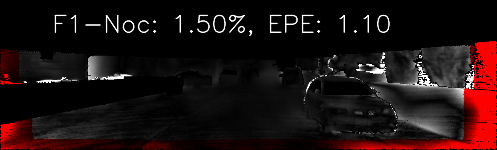} & 
\loadFig{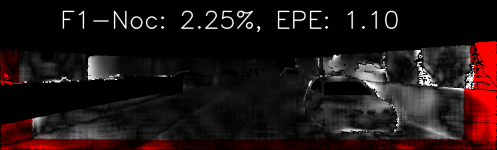} \\ 

\loadFig{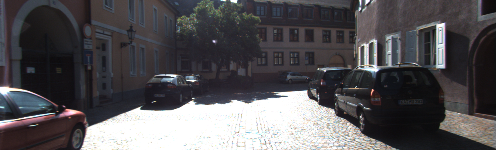} & 
\loadFig{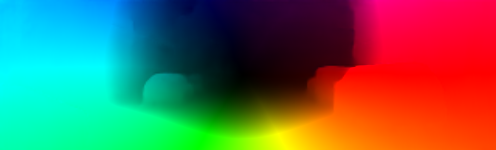} & 
\loadFig{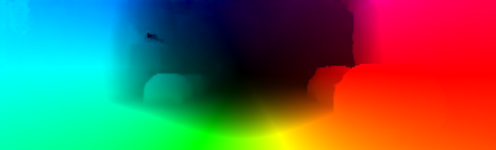} & 
\loadFig{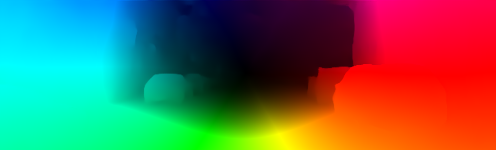} & 
\loadFig{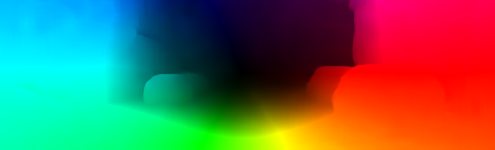} \\ 
\loadFig{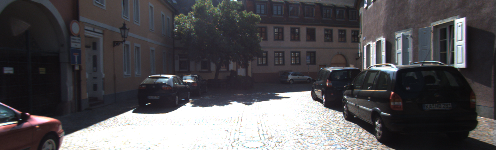} & 
\loadFig{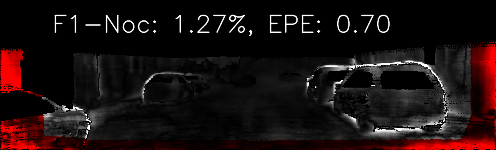} & 
\loadFig{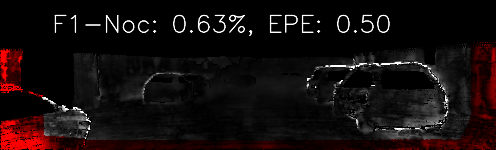} & 
\loadFig{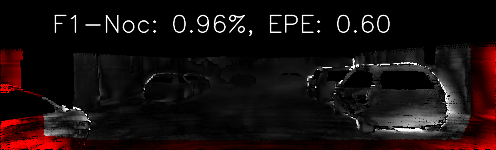} & 
\loadFig{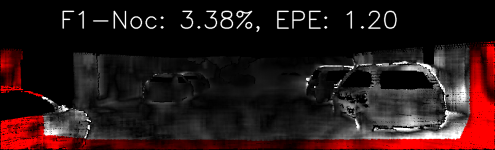} \\ 

\loadFig{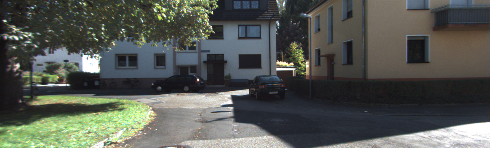} & 
\loadFig{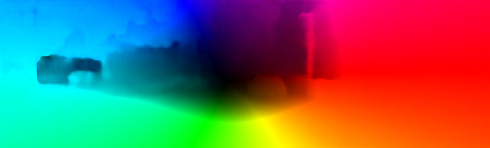} & 
\loadFig{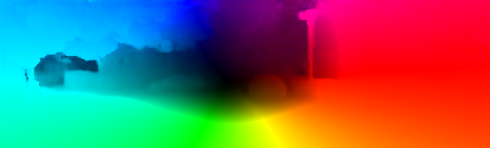} & 
\loadFig{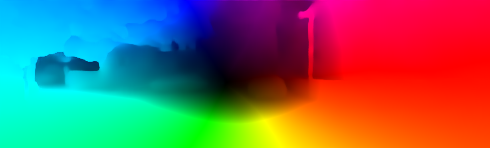} & 
\loadFig{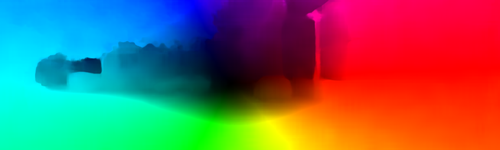} \\ 
\loadFig{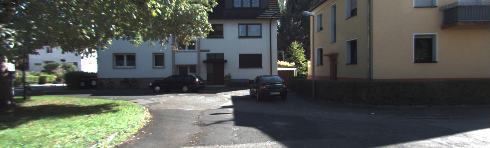} & 
\loadFig{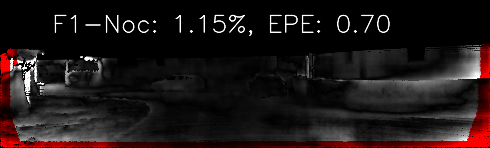} & 
\loadFig{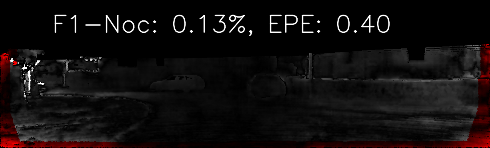} & 
\loadFig{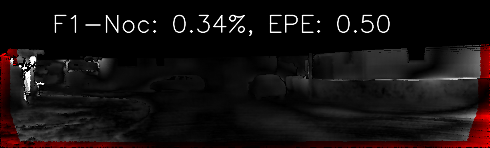} & 
\loadFig{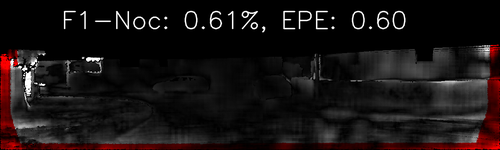} \\ 

\loadFig{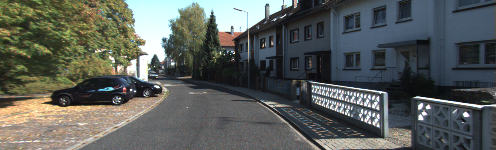} & 
\loadFig{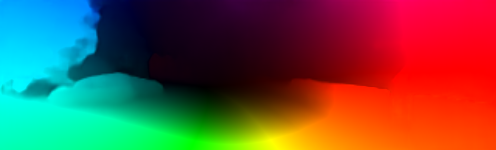} & 
\loadFig{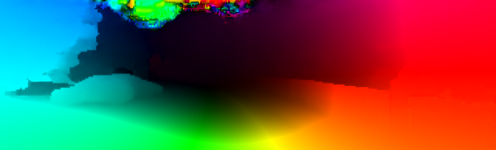} & 
\loadFig{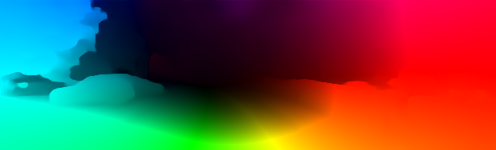} & 
\loadFig{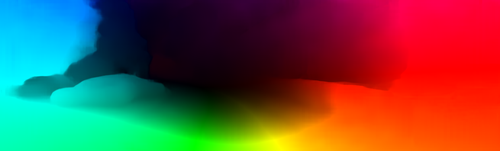} \\ 
\loadFig{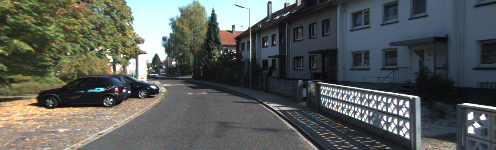} & 
\loadFig{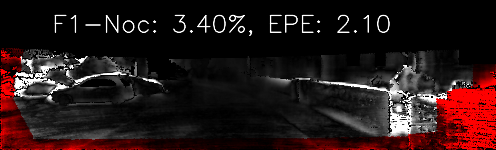} & 
\loadFig{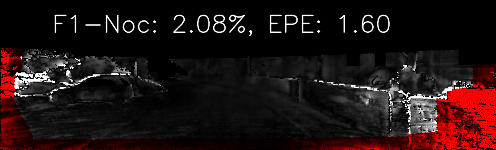} & 
\loadFig{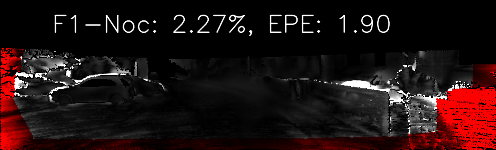} & 
\loadFig{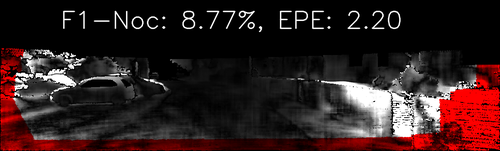} \\ 

\loadFig{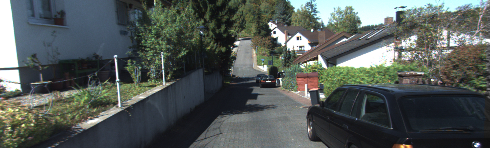} & 
\loadFig{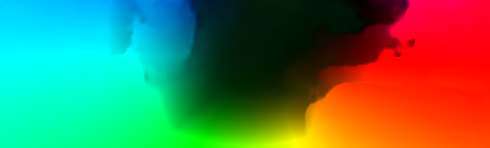} & 
\loadFig{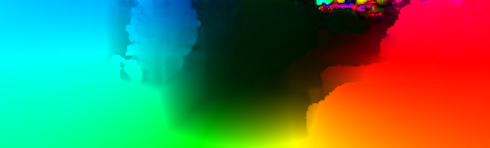} & 
\loadFig{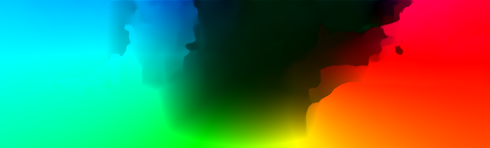} & 
\loadFig{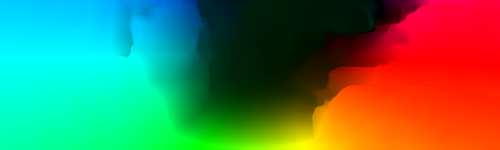} \\ 
\loadFig{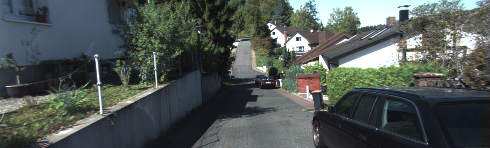} & 
\loadFig{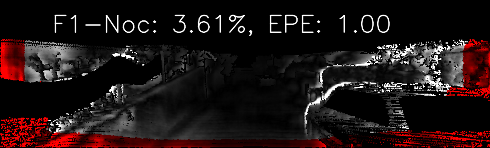} & 
\loadFig{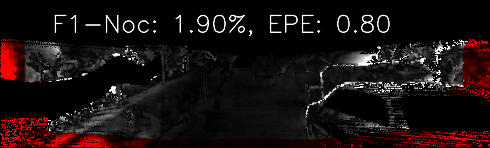} & 
\loadFig{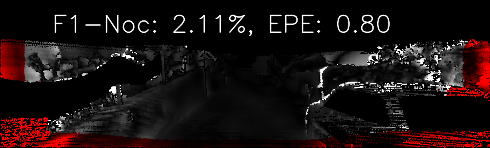} & 
\loadFig{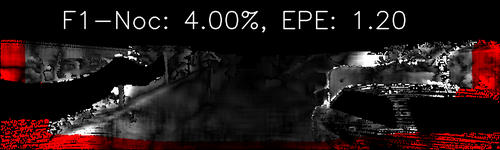} \\ 

\loadFig{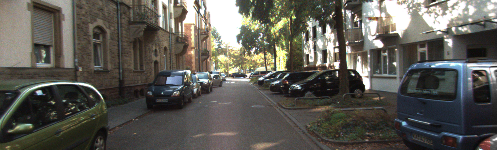} & 
\loadFig{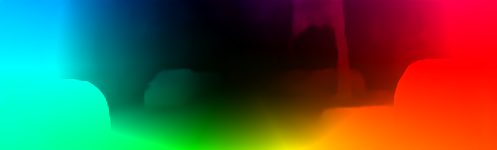} & 
\loadFig{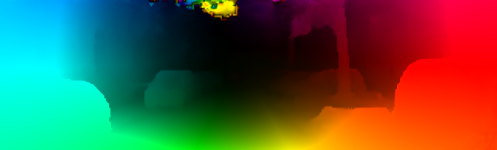} & 
\loadFig{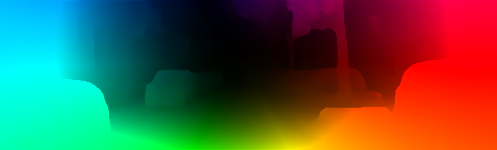} & 
\loadFig{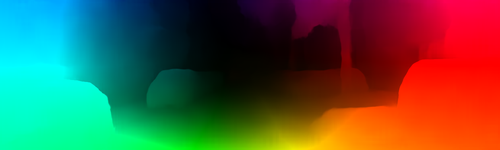} \\ 
\loadFig{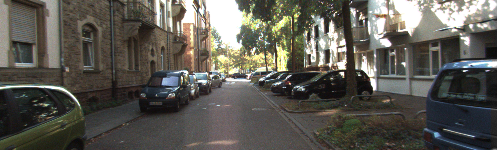} & 
\loadFig{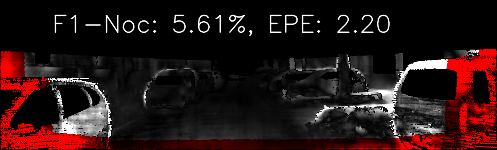} & 
\loadFig{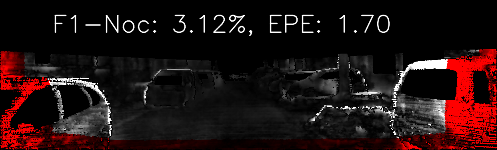} & 
\loadFig{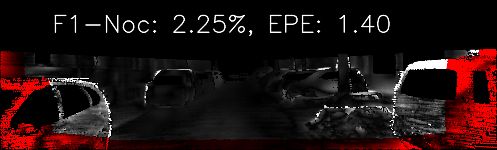} & 
\loadFig{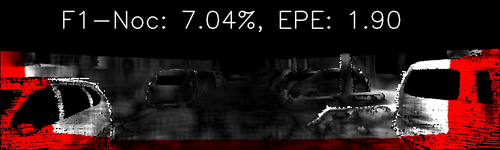} \\ 

\loadFig{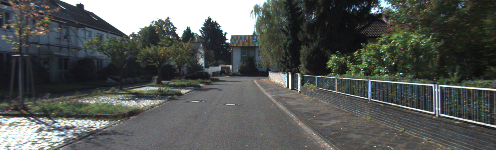} & 
\loadFig{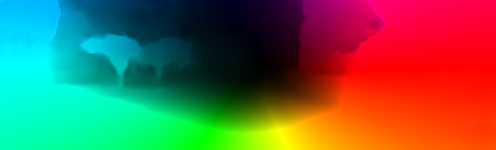} & 
\loadFig{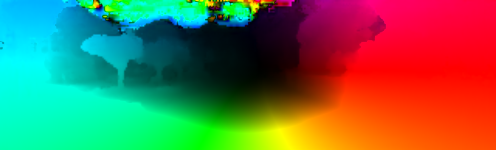} & 
\loadFig{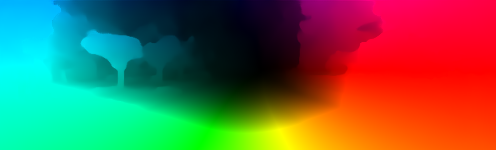} & 
\loadFig{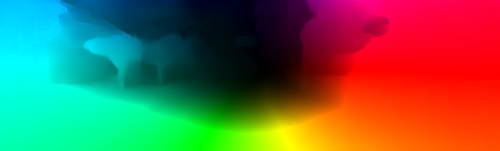} \\ 
\loadFig{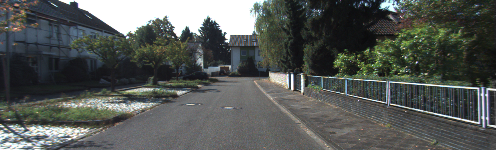} & 
\loadFig{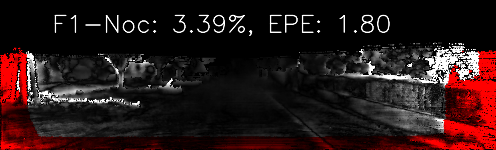} & 
\loadFig{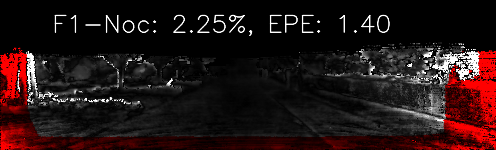} & 
\loadFig{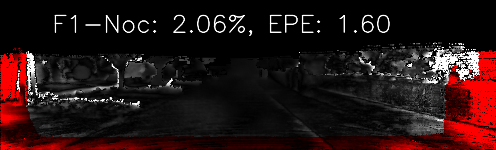} & 
\loadFig{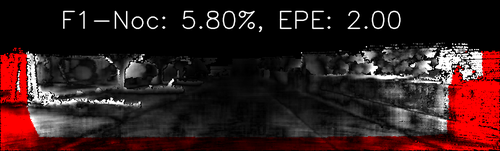} \\ 

(a) input images & (b) PWCNet~\cite{sun2018pwc} & (c) HD3~\cite{yin2019hierarchical} & (d) LiteFlowNet3~\cite{hui2020liteflownet3} & (e) our DCVNet \\
\end{tabular}
\caption{\textbf{Visual comparison of optical flow estimations on KITTI 2012}. For each method, we show colorized optical flow and error maps (obtained from online servers). For the error maps, white indicates large error while black means small error. Red denotes all occluded pixels, falling outside the image boundaries. Best viewed in color.}
\label{fig:res_k12}
\end{figure*}

We show more visual results of estimated optical flow from different approaches for MPI Sintel~\cite{Butler:ECCV:2012} in Fig.~\ref{fig:res_sintel_part_1} and Fig.~\ref{fig:res_sintel_part_2}. 
For KITTI 2015~\cite{Menze2018JPRS}, optical flow results are shown in Fig.~\ref{fig:res_k15_part_1} and Fig.~\ref{fig:res_k15_part_2}. 
The results of KITTI 2012 are shown in Fig.~\ref{fig:res_k12}. We also show optical flow results of our DCVNet on the DAVIS~\cite{Perazzi2016} dataset in Fig.~\ref{fig:res_davis} to demonstrate the cross-dataset generalization of our approach.

We made anonymous submission entries to all three benchmarks. We refer readers there for more visual results.

\newcommand{\AddImg}[1]{\includegraphics[width=0.33\textwidth]{#1}}
\begin{figure*}
\centering
\renewcommand{\tabcolsep}{0.25mm}
\begin{tabular}{ccc}
\AddImg{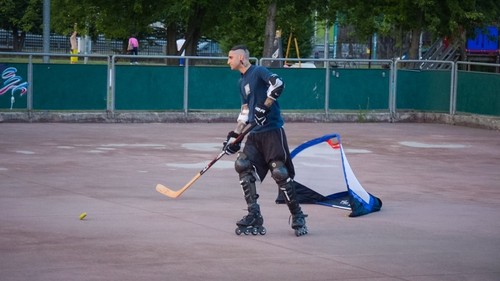} & 
\AddImg{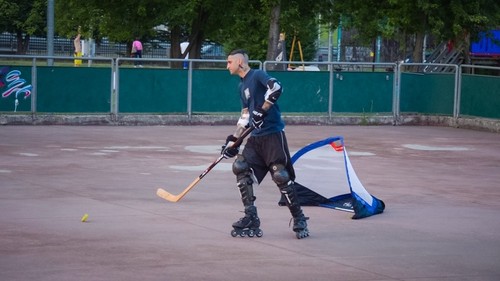} & 
\AddImg{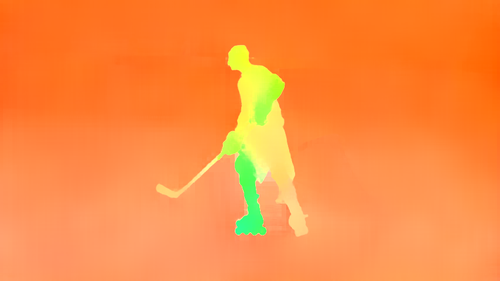} \\
\AddImg{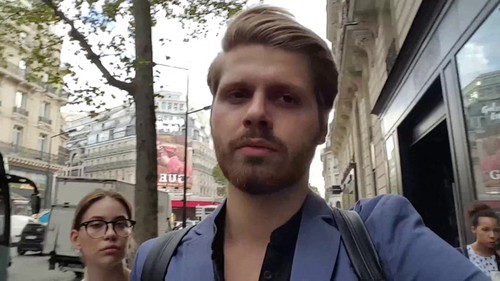} & 
\AddImg{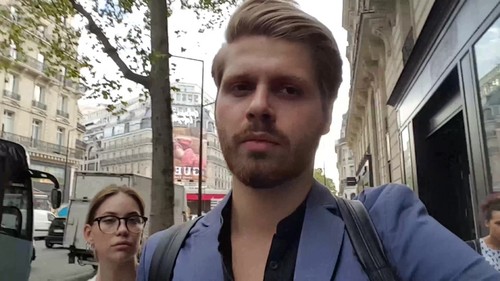} & 
\AddImg{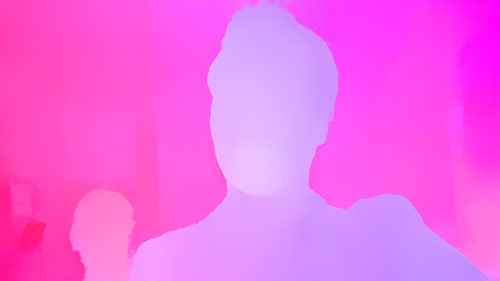} \\
\AddImg{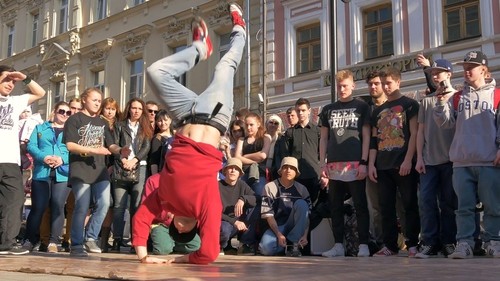} & 
\AddImg{} & 
\AddImg{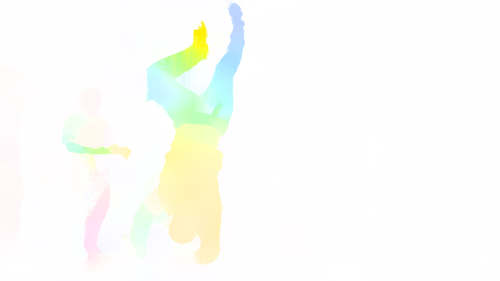} \\
\AddImg{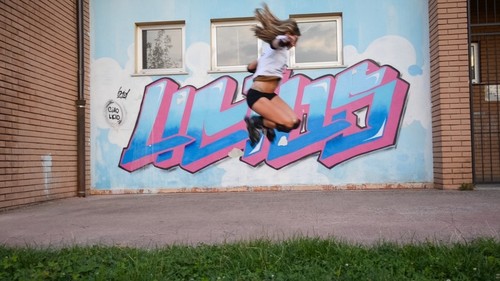} & 
\AddImg{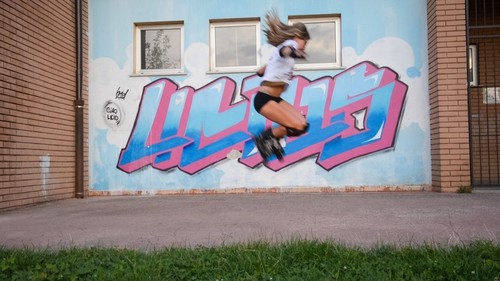} & 
\AddImg{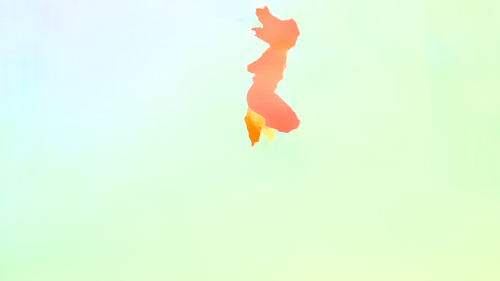} \\
\AddImg{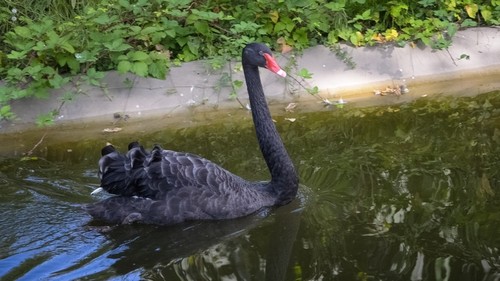} & 
\AddImg{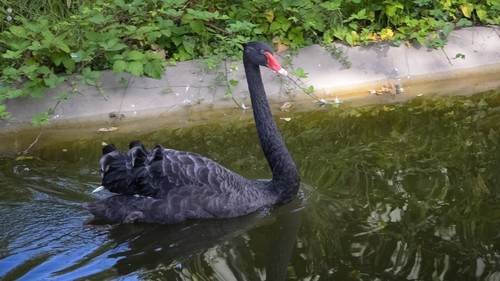} & 
\AddImg{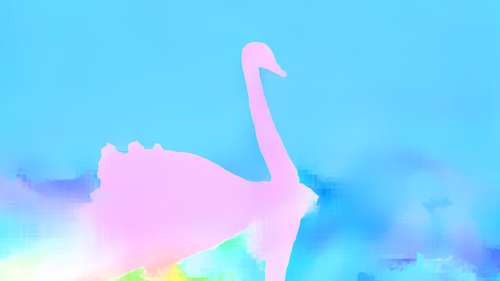} \\
\AddImg{} & 
\AddImg{} & 
\AddImg{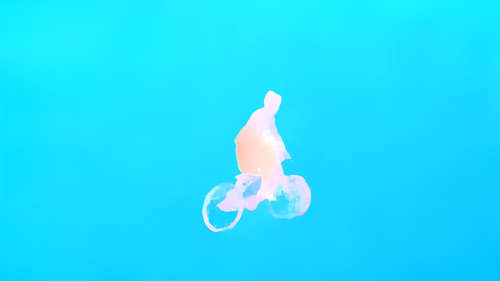} \\
(a) & (b) & (c) \\
\end{tabular}
\caption{\textbf{Visual results of estimated optical flow for the DAVIS~\cite{Perazzi2016} dataset.} From left to right: (a) first input images, (b) second input images, and (c) estimated optical flow. The last two rows are failure cases.}
\label{fig:res_davis}
\end{figure*}
\end{appendices}

\end{document}